%% file: NeuralNetworks2021.tex
\documentclass{article}
\usepackage{arxiv}
\usepackage{hyperref}
\usepackage{url}
\usepackage{doi}
\usepackage{graphicx}
\usepackage{amssymb, amsmath}
\usepackage{multirow}
\usepackage{subfigure}
\usepackage{latexsym}
\usepackage[symbol]{footmisc}
\usepackage{wrapfig}
\usepackage{caption}
\usepackage[round]{natbib}

\usepackage{color}
\usepackage{fancyhdr}

\title{{\bf \fontsize{20pt}{0pt}\selectfont Sensitivity \\ \vspace{1mm}
\fontsize{16pt}{0pt}\selectfont  -- Local Index to Control Chaoticity or Gradient Globally -- \footnotemark[1]}}

\author{Katsunari Shibata\footnotemark[2]\ \ \ \ \ \ 
Takuya Ejima\ \ \ \ \ \ 
Yuki Tokumaru\footnotemark[3]\ \ \ \ \ \ 
Toshitaka Matsuki\\ \\ 
\texttt{katsunarishibata@gmail.com, matsuki@oita-u.ac.jp} \\ \\
Oita University, Japan}

\begin{document}
\maketitle

\begin{abstract}
Here, we introduce a fully local index named ``sensitivity'' for each neuron to control chaoticity or gradient globally in a neural network (NN). We also propose a learning method to adjust it named ``sensitivity adjustment learning (SAL)''. The index is the gradient magnitude of its output with respect to its inputs.
By adjusting its time average to 1.0 in each neuron, information transmission in the neuron changes to be moderate without shrinking or expanding for both forward and backward computations. That results in moderate information transmission through a layer of neurons when the weights and inputs are random. Therefore, SAL can control the chaoticity of the network dynamics in a recurrent NN (RNN). It can also solve the vanishing gradient problem in error backpropagation (BP) learning in a deep feedforward NN or an RNN. We demonstrate that when applying SAL to an RNN with small and random initial weights, log-sensitivity, which is the logarithm of RMS (root mean square) sensitivity over all the neurons, is equivalent to the maximum Lyapunov exponent until it reaches 0.0. We also show that SAL works with BP or BPTT (BP through time) to avoid the vanishing gradient problem in a 300-layer NN or an RNN that learns a problem with a lag of 300 steps between the first input and the output. Compared with manually fine-tuning the spectral radius of the weight matrix before learning, SAL's continuous nonlinear learning nature prevents loss of sensitivities during learning, resulting in a significant improvement in learning performance.

$\langle$ Highlights $\rangle$
\begin{itemize}
\item ``Sensitivity" is a local version of the maximum Lyapunov exponent for each neuron.
\item Log sensitivity is equivalent to the maximum Lyapunov exponent until the dynamics reach the ``edge of chaos".
\item Sensitivity Adjustment Learning (SAL) adjusts the sensitivity in each neuron and realizes ``edge of chaos" in the network as a result.
\item SAL also prevents ``vanishing gradient" in gradient-based learning such as BP or BPTT, which greatly improves learning performance.
\item Compared with the adjustment of weight matrix, SAL can consider non-linearity and prevent the loss of sensitivity caused by another learning.
\end{itemize}
\end{abstract}


\pagestyle{fancy}
\title{Sensitivity\hspace{1mm}
{\small  -- Local Index to Control Chaoticity and Gradient Globally --
\hspace{9mm}
  K.~Shibata, T.~Ejima, Y.~Tokumaru, T.~Matsuki}}

\footnotetext[1]{Published in {\it Neural Networks}, vol.~143, pp.~436--451 (2021)\ \ \ \ 
\url{https://doi.org/10.1016/j.neunet.2021.06.015} }
\footnotetext[2]{Corresponding author, No current affiliation}
\footnotetext[3]{Current affiliation is NEC Solution Innovators, Ltd.}

\section{Introduction}
Deep learning using a deep feedforward neural network (DFNN) or a recurrent neural network (RNN)
has attracted great attention due to its drastic performance improvement especially in recognition of patterns
including the case of time-series signals~[\cite{SurveyObjectDetection, SurveySpeechRecog, SurveyTimeSeries, SurveyLangProc}].
They have shown us the overwhelming power of massively parallel processing systems
acquired through learning.
That suggests the way towards human-like processing,
which human designers have struggled to design for a long time,
but has not been achieved by conventional artificial intelligence.

Since higher functions such as thinking and communication cannot be discussed 
without processing in the time axis,
there is no doubt that the significance of temporal processing or dynamics will increase
more and more, and longer processing will be required from now on.
In temporal processing, we need to think of the internal state of a network
not as points in space but as lines or flow formed around a point moving along time.

When solving a task with long-term dependency is required,
the effective preservation of information in a neural network through time
becomes critical. 
That is deeply related to the chaoticity of the network.
If the maximum Lyapunov exponent of the network is negative,
much of the information shrinks and disappears over time.
If positive, the network expands even trivial pieces of information one after another,
but in contrast, it cannot magnify rather large pieces of information so much
due to the nonlinearity by the limited value range in each neuron.
Therefore, even though they were large, it is challenging to retrieve the old information
from the current internal state after a long time lag.
Accordingly, for effective information preservation through time,
dynamics around the ``edge of chaos'' would be the right choice.

As for the learning of long-term dependency in an RNN,
vanishing/exploding gradient in BPTT (error Back Propagation Through Time)
as a gradient-based learning method has been discussed
since far before the boom of deep learning~[\cite{Bengio94, Hochreiter98, DifficultRNN}].
This problem is caused by the same logic as the information preservation
for a long time lag mentioned above.
The gradient here is the gradient of some evaluation (can be cost or error) function of final outputs
with respect to the weight vector or neuron state vector.
By decomposing the gradient by chain rule,
we can understand the problem comes from the repetition
of the information scaling through each time-step.
A similar discussion can be made for the same problem in deep feedforward NNs (DFNNs).
Detailed and related works are described in the next section.

Reservoir computing is often used to learn tasks that need temporal processing
and shows excellent results
even though it usually does not learn the connection weights among the reservoir neurons.
Here, an important parameter is the scale or spectral radius of the weight matrix in the reserver.
Echo state property is considered so that the effect of initial conditions vanish
as time passes~[\cite{Re-visitingESP}].
For learning long-term dependency, the spectral radius should be close to 1.0.
In FORCE learning, the learning performance is good when the network dynamics are
around the edge of chaos~[\cite{FORCE}].
We have also shown in reward-modulated Hebbian learning with chaotic exploration
using a reservoir network,
dynamics around the edge of chaos brings out good learning performance~[\cite{Matsuki, NN_Matsuki}].
However, the reservoir does not learn its dynamics directly.
That is similar to the perceptron that fixes its randomly-decide hidden weights and does not learn
its hidden representation.
Furthermore, we cannot find how such random weights with an appropriate size
are realized in each neuron autonomously without any centralized system.
We believe that learning dynamics must be crucial
when developing higher functions like ``thinking''.

To develop a ``thinking machine'' as an ultimate artificial intelligence system,
autonomous and rational state transitions even without the help of external stimuli must be essential
in its RNN.
Similar to the other functions such as ``recognition'' and ``memory'',
we expect ``thinking'' emerges within the framework of end-to-end reinforcement learning
using an RNN~[\cite{End-to-End}].
Acquiring appropriate memories that do not need transitions but need convergence
is relatively easy for an agent~[\cite{AROBJ04, Utsunomiya, Utsunomiya2, Prediction}].
However, it is not easy to acquire rational transitions through reinforcement learning using a regular RNN
even though the transitions are externally driven~[\cite{Sawatsubashi}].

Then, we focused on chaotic dynamics generated in an RNN
and proposed a new reinforcement learning (RL) paradigm~[\cite{IJCNN2015, Chaos-RL}].
Chaotic dynamics expand tiny variations in its network state through time.
That destabilizes the dynamics and enables autonomous state transition in the RNN,
although the transition is irregular.
In the new proposed RL, exploration does not use stochastic action selection by random noises
but uses the autonomous state transition based on the internal chaotic dynamics.
We expect that the irregular dynamics would become rational
by forming attractors on the dynamics through learning.
According to the above discussion, we set up a hypothesis
that ``exploration'' grows into ``thinking'' through learning.
In our new RL, the chaoticity of the RNN also should not be either
too strong or too weak~[\cite{Goto, Sato}].
Adjustment of chaoticity is critical in this learning paradigm as well.

Processing in each neuron produces the dynamics of the network.
The propagated error signals in gradient-based BP (error backpropagation) or BPTT learning
represent the influence of tiny variation in each neuron's state on the final error function.
Therefore, we expect that by adjusting the influence in each neuron,
we can control both chaoticity and error backpropagation of the network simultaneously.
In this paper, we define the gradient magnitude of its output
with respect to its input vector as ``sensitivity'' in each neuron
and use it as a local index to control the chaoticity and error signal propagation globally
in the network.
We also propose a learning method to adjust the sensitivity based on hill-climbing
and call it ``sensitivity adjustment learning (SAL)''.
Then we show that an RNN comes to generate chaotic dynamics through SAL,
and observe the relationship between the sensitivities as a local index
and the maximum Lyapunov exponent as a global index during learning.
We also show some supervised learning results of a simple problem
with a lag of 300 steps between the first input and final output timings
when using both BPTT and SAL in an RNN, and the learning process is analyzed.
Finally, we apply SAL to DFNNs with BP and show a 300-layer DFNN can learn a simple problem
with noise addition and observe the processing.

\section{Related Works}
Several techniques have been proposed already to avoid the vanishing/exploding gradient problem
in error backpropagation (BP) or backpropagation through time (BPTT).
They can be roughly divided into two categories.

In the first category, which mainly targets on RNNs whose feedback weight matrix is square,
the spectral radius of the Jacobian matrix between the inputs and outputs of a layer or group of neurons
is directly set to be around 1.0.
The simplest way is to set the Jacobian matrix close to the identity matrix.
In this case, the output vector is close to the input vector.
Therefore, a slight variation in each neuron state does not change so much through the forward processing.
The error signal in each neuron also does not change so much through the backward processing.

LSTM~[\cite{LSTM}], a quite popular RNN, employs special units named LSTM cells.
If the forget gate is fully open, the cell state does not change, and
each diagonal element of the Jacobian matrix is close to 1.0
when only the signal flow inside the cell is focused on.
GRU~[\cite{GRU}] has a simpler but similar structure to LSTM.
We have adopted a far simpler method using a regular RNN
with setting all the self-feedback connection weights
to the reciprocal of the maximum derivative of the activation function.
Concretely, the weight value is 4.0 for the sigmoid function and 1.0 for the hyperbolic tangent function.
All the other feedback connection weights are set to 0.0 or small random values.
In this case, the Jacobian matrix is close to the identity matrix when the activations are small enough
in the activation function's linear region.
It works well in learning memory-required tasks~[\cite{AROBJ04, Utsunomiya, Utsunomiya2, Prediction}],
which need to form fixed-point attractors or static associative memory.
In DFNNs, shortcut connections through one or more layers as in a ResNet~[\cite{ResNet}],
which is widely used mainly in convolutional NNs (CNNs),
also make its Jacobian matrix close to the identity matrix if the other connection weights are small.

However, the transformation represented by the identity matrix
is equivalent to applying no processing.
Therefore, it is suitable to keep some information without any change,
but it is not appropriate to learn a complicated conversion of input signals or
internal dynamic state transition like ``thinking''.
Actually, in our work~[\cite{Sawatsubashi}],
although an agent could learn simple state transitions in this approach,
learning was so difficult that careful design of task shaping was necessary.

In the second category, the mean and variance of neuron activations are normalized,
usually to zero mean and unit variance.
Batch normalization~[\cite{BatchNormalization}], layer normalization~[\cite{LayerNormalization}],
weight normalization~[\cite{WeightNormalization}], self-normalizing neural networks~[\cite{SNNN}]
can be categorized here in a broad meaning.
In the self-normalizing neural network, by setting the weight matrix and activation function appropriately,
the activations close to zero mean and unit variance converge towards zero mean and unit variance.
Furthermore, the variance of neuron activations is bounded, and the network does not suffer
from a vanishing gradient.
In the weight normalization process, the weight vectors are initialized depending on the given data
so that the mean and variance of neuron activations are normalized.

Different from the approaches mentioned above, moderatism aims to acquire necessary processing
by keeping the variation of both inputs and output in each neuron moderate~[\cite{Moderatism}].
Here, we focus on the sensitivity that represents magnification or contraction of a small variation
in each neuron's processing.
By controlling it in each neuron, the network's global dynamics can be controlled.
We show that adjusting each neuron's sensitivity enables learning for DFNNs
or RNNs solving a long time-lag problem.

On the other hand, from the viewpoint of chaos control, previous studies have mainly focused on
stabilizing a system with chaotic dynamics~[\cite{OGY, Control_Chaos}] and have not positively utilized
the chaos or edge of chaos dynamics.
From the information storage or processing perspective, the importance of the ``edge of chaos''
has been highlighted by several studies~[\cite{Langton, Legenstein, Boedecker}].
Moreover, from the biological viewpoint, the relation between known input stimuli and attractors
in brain dynamics was investigated, and the role of the chaotic dynamics for new input stimuli
was addressed~[\cite{Skarda, Freeman}].
Based on the study, an associative memory model using a chaotic neural network
was also proposed~[\cite{Osana}].
However, it is not easy to find a method to generate and control chaotic dynamics
by learning in an RNN.
Consequently, dynamics around the edge of chaos produced in an RNN with fixed weights
have been utilized widely as a reservoir as mentioned above.
In this paper, aiming to positively utilize chaos or edge of chaos dynamics,
we propose to generate and control chaotic dynamics in RNNs by adjusting local sensitivity
in each neuron.

\input{Theory}

\input{Simulations}

\section*{Acknowledgment}
The authors would like to thank Prof.~Hiromichi Suetani for useful suggestions about chaotic dynamics.
They also thank their laboratory members for discussions and sharing related simulation results.
This work was supported by JSPS KAKENHI Grant Number JP15K00360, JP20K11993
and also Kayamori Foundation of Informational Science Advancement K31-KEN-XXIV-539.
\newpage
\bibliographystyle{abbrvnat}
\bibliography{Reference}
\end{document}

%% file: Theory.tex
\section{Sensitivity and Sensitivity Adjustment Learning (SAL)}
\subsection{Definition of Sensitivity}
Fig.~\ref{fig:neuron_forward} shows a general static-type neuron model with $m$ inputs.
Its internal state $u$ is derived as the inner product of the input vector
${\mathbf{x}}=(x_1, \ldots  , x_m)^{\mathrm{T}}$ and connection weight vector
${\mathbf{w}}=(w_1, \ldots  , w_m)^{\mathrm{T}}$ as
\begin{equation}
u = {\mathbf{w}}\cdot{\mathbf{x}},
\label{Eq:internal state}
\end{equation}
and the output $o$ is derived as
\begin{equation}
o = f(U)=f(u+\theta),
\label{Eq:output}
\end{equation}
where $U=u+\theta$, $\theta$ is the bias, and $f(\cdot)$ is an activation function that can be hyperbolic tangent or sigmoid function.
The sensitivity introduced here is a local index for each neuron to show
how the neuron is sensitive to a small change in its inputs.
It is defined as the Euclidean norm of the output gradient with respect to the input vector ${\mathbf x}$ as
\begin{equation}
s(U; {\mathbf{w}}) = \left\|\nabla_{\mathbf{x}} o\right\| = f'(U)\|{\mathbf{w}}\|
\label{Eq:sensitivity}
\end{equation}
if the activation function $f(\cdot)$ is a monotonically increasing function.
If the vector elements are used, it is rewritten as
\begin{equation}
s(U; {\mathbf w}) = \sqrt{\sum_i^m \left( \frac{\partial o}{\partial x_i} \right)^2} = f'(U)\sqrt{\sum_i^m w_i^2}\ .
\label{Eq:sensitivity_non_vector}
\end{equation}

\subsection{Sensitivity in  the forward computation}
\begin{figure}[b]
\centerline{\includegraphics[scale=0.6]{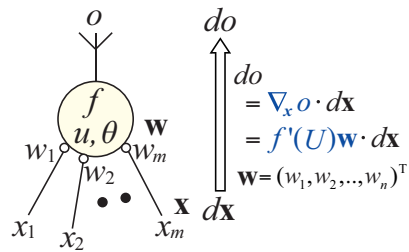}}
\caption{A neuron model and the influence of a small deviation in its inputs ${\mathbf x}$ on its output $o$.}
\label{fig:neuron_forward}
\end{figure}

In the forward computation in the neuron, as shown in Fig.~\ref{fig:neuron_forward},
an infinitesimal change in the inputs $d{\mathbf x}$ produces the infinitesimal change
in the output $do$ as 
\begin{equation}
do = \nabla_{\mathbf x} o \cdot d{\mathbf x} = f'(U){\mathbf w} \cdot d{\mathbf x},
\end{equation}
and can be rewritten as
\begin{equation}
do = \left\|\nabla_{\mathbf x} o \right\| \left\|d{\mathbf x}\right\| \cos\phi = s(U; {\mathbf w}) \left\|d{\mathbf x}\right\| \cos\phi
\label{Eq:derivative}
\end{equation}
using the direction cosine $\cos \phi$ between the input change vector $d{\mathbf x}$
and the weight vector ${\mathbf w}$ whose direction is the same as the gradient vector $\nabla_{\mathbf x} o$.

\begin{figure}[t]
\centerline{\includegraphics[scale=0.9]{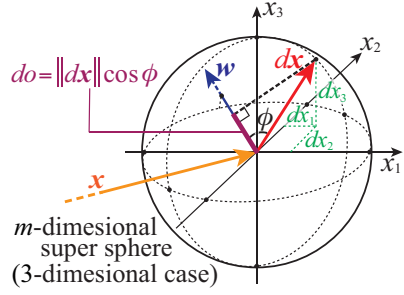}}
\caption{The relation between an infinitesimal input change vector $d{\mathbf x}$ and 
the output change $do$ in the case of  the number of inputs $m=3$.
When the sensitivity $s$ is 1.0, $do$ is the projection of $d{\mathbf x}$ onto the weight vector ${\mathbf w}$
whose direction is the same as $\nabla_{\mathbf x} o$.
Therefore, when $d{\mathbf x}$ is distributed uniformly on a super sphere surface,
the variance of $do$ is equivalent to that of one of the $m$ input signals.}
\label{fig:supersphere}
\end{figure}
Here, $d{\mathbf x}$ is assumed to be an $m$-dimensional standard normal random vector
multiplied by an infinitesimal constant $\epsilon$
as $d{\mathbf x} \sim \mathcal{N}(0,\epsilon^2 I_m)$ where $I_m$ is the $m$-dimensional unit matrix.
The distribution of the vector direction is uniform, and that is the same as
the uniform distribution on the $m$-dimensional super sphere surface
as presented in Fig.~\ref{fig:supersphere}.
In an $m$-dimensional Euclidean space, the square of the vector $d{\mathbf x}$ is identical to
the sum of its $m$ squared elements.
In other words, the sum of the $m$ squared direction cosine to individual standard bases is 1.0.
Therefore, from the symmetry among $m$ dimensions, the square of the projection of $d{\mathbf x}$
on any direction is expected to be $1/m$ of the square of $d{\mathbf x}$ itself.
By applying it to the relation between the vector $d{\mathbf x}$ and its elements $dx_i$,
the square of the vector $d{\mathbf x}$ is expected to be the $m$-times of the variance
of each element $dx_i$ as
\begin{equation}
\mathit{E}\left [\left\| d{\mathbf x} \right\|^2 \right ] = \sum_{i}^{m} \mathit{V}\left[dx_i \right] = m \mathit{V}\left[dx_i \right]=m\epsilon^2.
\label{Eq:E_dx2}
\end{equation}
By applying it to the projection of the vector $d{\mathbf x}$ onto the weight vector ${\mathbf w}$,
the variance of the direction cosine $\cos \phi$ is derived as
\begin{equation}
\mathit{V}\left[\cos \phi \right] = \mathit{E}\left [\cos^2 \phi \right ] = \frac{1}{m}
\label{Eq:V_cos_phi}
\end{equation}
where the expected direction cosine is 0.0 from the symmetry of the $d{\mathbf x}$ distribution.
The direction of $d{\mathbf x}$ is uniformly distributed
and independent with the size of $\|d{\mathbf x}\|$.
Then, in Eq.~(\ref{Eq:derivative}), $\left\|d{\mathbf x}\right\|$ and $\cos\phi$ are independent.
Therefore, from Eqs.~(\ref{Eq:E_dx2}) and (\ref{Eq:V_cos_phi}),
the variance of the output change $do$ can be written for a given sensitivity $s$
using the variance of one input change $dx_i$ as
\begin{equation}
\mathit{V}\left[do \right] = \{s(U; {\mathbf w})\}^2\mathit{E}\left [\left\| d{\mathbf x} \right\|^2 \right ]
    \mathit{E}\left [ cos^2 \phi \right ] = s^2\mathit{V}\left[dx_i \right]=s^2\epsilon^2.
\label{Eq:projected_var}
\end{equation}
If the sensitivity $s$ is 1.0, then the output change in Eq.~(\ref{Eq:derivative}) can be written as
\begin{equation}
do =\left\|d{\mathbf x}\right\| \cos\phi,
\label{Eq:Ratio}
\end{equation}
and that is the projection of the $m$-dimensional infinitesimal vector $d{\mathbf x}$ onto the direction
of the weight vector ${\mathbf w}$ as shown in Fig.~\ref{fig:supersphere}.
Therefore, its distribution becomes $do \sim \mathcal{N}(0,\epsilon^2)$ not depending on the number of inputs $m$, and Eq.~(\ref{Eq:projected_var}) is changed as follows:
\begin{equation}
\mathit{V}\left[do \right] = \mathit{V}\left[dx_i \right]=\epsilon^2.
\end{equation}
This is significant because the neuron maintains the variance of just one input signal
as its own output variance without being shrunk or expanded
not depending on the number of connections $m$.

\begin{figure}[bht]
\centerline{\includegraphics[scale=0.8]{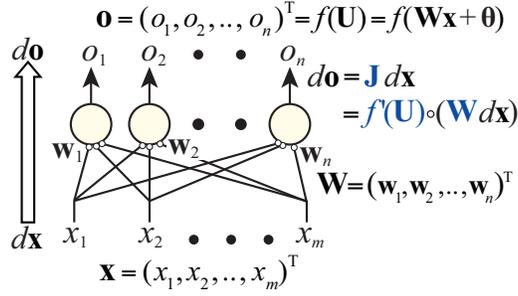}}
\caption{Forward (output) computation through a layer of neurons.}
\label{fig:layer_forward}
\end{figure}
Next, the case of a group or layer of $n$ neurons as can be seen in Fig.~\ref{fig:layer_forward} is considered,
and its output vector is indicated as ${\mathbf o}=(o_1, \ldots , o_n)^\mathrm{T}$.
The infinitesimal output change $d{\mathbf o}$ can be represented using the infinitesimal input change $d{\mathbf x}$
as
\begin{equation}
d{\mathbf o} = {\mathbf J}({\mathbf{U}};{\mathbf{W}}) d{\mathbf x}=f'({\mathbf U})\circ({\mathbf W}d{\mathbf x})
\label{Eq:Layer}
\end{equation}
where $f'({\mathbf U})=(f'(U_1), \ldots ,f'(U_n))^\mathrm{T}$ and
`$\circ$' indicates element-wise multiplication.
${\mathbf W}=({\mathbf w}_1, \ldots ,{\mathbf w}_n)^\mathrm{T}$ is an $n \times m$ weight matrix
where ${\mathbf w}_j (j=1, \ldots ,n)$ is the weight vector of the $j$th neuron.
${\mathbf J}({\mathbf U}; {\mathbf W})$ is the Jacobian matrix as
\begin{equation}
 {\mathbf J}({\mathbf U}; {\mathbf W}) = \begin{pmatrix}
 \frac{\partial o_1}{\partial x_1} & \cdots & \frac{\partial o_1}{\partial x_m} \\
 \vdots & \ddots & \vdots \\
 \frac{\partial o_n}{\partial x_1} & \cdots & \frac{\partial o_n}{\partial x_m} 
\end{pmatrix}
= \begin{pmatrix}
(\nabla_{\mathbf x} o_1)^\mathrm{T} \\
\vdots \\
(\nabla_{\mathbf x} o_n)^\mathrm{T}
\end{pmatrix}
= \begin{pmatrix}
f'(U_1){\mathbf w}_1^\mathrm{T} \\
\vdots \\
f'(U_n){\mathbf w}_n^\mathrm{T}
\end{pmatrix}.
\label{Eq:JacobianMatrix}
\end{equation}
Eq.~(\ref{Eq:Layer}) can be rewritten using sensitivities of the neurons as
\begin{equation}
d{\mathbf o} = \begin{pmatrix}
\nabla_{\mathbf x} o_1 \cdot d{\mathbf x}\\
\vdots \\
\nabla_{\mathbf x} o_n \cdot d{\mathbf x}
\end{pmatrix}
= \begin{pmatrix}
s_1 \left\|d{\mathbf x}\right\| \cos \phi_1\\
\vdots \\
s_n \left\|d{\mathbf x}\right\| \cos \phi_n
\end{pmatrix}.
\end{equation}
Here, in addition to the assumption of $d{\mathbf x}$ mentioned above,
it is further assumed that the weight matrix ${\mathbf W}$ has been randomly chosen
with i.i.d.~elements with mean 0,
and also $f'({\mathbf U})$ is a random vector having elements with an identical distribution.
If the number of output $n$ is not greater than the number of input $m$,
the elements of the infinitesimal change of the output vector $do_j$ are i.i.d.~with mean 0.
The mean square of the magnitude of the vector $d{\mathbf o}$ is $n$ times the
variance of one output change $do_j$ as
\begin{equation}
\mathit{E}\left [\left\| d{\mathbf o} \right\|^2 \right ] = \sum_j^n \mathrm{V}[do_j] = n\mathrm{V}[do_j].
\end{equation}
From Eqs.~(\ref{Eq:V_cos_phi}) and (\ref{Eq:projected_var}), 
\begin{equation}
\frac{1}{n}\mathit{E}\left [\left\| d{\mathbf o} \right\|^2 \right ]
    = \frac{\langle s^2 \rangle}{m}\mathit{E}\left [\left\| d{\mathbf x} \right\|^2 \right ]
    =\langle s^2 \rangle \epsilon^2
\label{Eq:layer_do_dx}
\end{equation}
on the assumption that $n$ is large enough and
\begin{equation}
\langle s^2 \rangle = \mathit{E}\left [\{s(U_j; {\mathbf w_j})\}^2 \right] = \mathit{E}\left [\{f'(U_j){\mathbf w}_j\}^2 \right]
\ \ \ (j=1,\dots,n)
\end{equation}
where $\langle s^2 \rangle$ is the mean of $s^2$ over the $n$ neurons.
If $\langle s^2 \rangle =1.0$ or the sensitivity of each neuron is 1.0, then
\begin{equation}
\frac{\mathrm{RMS}\left [\left\| d{\mathbf o} \right\| \right ] }{\sqrt{n}}
    = \frac{\mathrm{RMS}\left [\left\| d{\mathbf x} \right\| \right ] }{\sqrt{m}} = \epsilon
\end{equation}
where $\mathrm{RMS}$ means root mean square.
If the numbers of inputs $m$ and outputs $n$ are the same, then
\begin{equation}
\mathrm{RMS}\left [\left\| d{\mathbf o} \right\| \right ] = \mathrm{RMS}\left [\left\| d{\mathbf x} \right\| \right ]
= \sqrt{m}\epsilon
\end{equation}
When the number of neurons $n$ is less than the number of inputs $m$,
it is expected that the amount of information is decreased through the layer.

From the above discussion, if each neuron's sensitivity is 1.0,
the distribution of the small change in each neuron is maintained
not depending on the number of neurons or number of inputs.
Accordingly, a small change in an input signal of the network reaches the final network output
even though (1) the number of layers is large,
(2) the number of neurons in each layer is varied
and/or (3) the connection between layers is sparse.
The same discussion can be applied to an RNN.
The distribution of small output change is maintained through time
regardless of the structure of the RNN,
such as sparse or full connections, flat (non-layered) or layered.
Therefore, if the sensitivities for all the neurons are around 1.0,
the maximum Lyapunov exponent as the logarithm of time development of small change
is expected to be around 0.0.

\subsection{Sensitivity in the backward computation}
\begin{figure}[t]
\centerline{\includegraphics[scale=0.6]{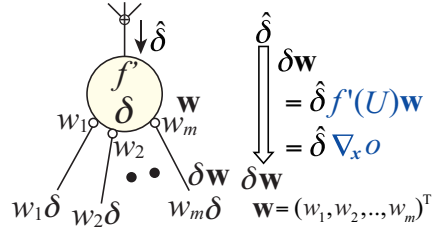}}
\caption{Backward (error signal) computation in a neuron.}
\label{fig:neuron_backward}
\end{figure}

As mentioned in the Introduction, when a small change in network inputs reaches the network outputs,
the error signals for learning reach the input layer from the output layer
in the backward computation for gradient-based learning such as BP or BPTT.
A neuron receives an error signal for its output $\hat{\delta} = -\partial E/\partial o$
where $E$ is a given cost or error function.
Error signal $\delta$ for its internal state $u$ is derived by multiplying the derivative of its activation function
to $\hat{\delta}$ as
\begin{equation}
\delta = -\frac{\partial E}{\partial u} =  -\frac{\partial E}{\partial o} \frac{do}{du} = \hat{\delta}f'(U).
\label{Eq:multi_derivative}
\end{equation}
As depicted in Fig.~\ref{fig:neuron_backward}, the error signal is propagated after being weighted
by the weight vector ${\mathbf w}$ before propagating to the one-level lower-layer neurons.
Then the propagated error signal vector $\delta {\mathbf w}$ can be written as
\begin{equation}
\delta {\mathbf w} = \hat{\delta}f'(U){\mathbf w} = \hat{\delta}\nabla_{\mathbf x} o.
\end{equation}
Therefore, its magnitude can be expressed as
\begin{equation}
\left\|\delta {\mathbf w} \right\| = \left\| \hat{\delta}\nabla_{\mathbf x} o \right\| 
= \left| \hat{\delta}\right|\left\|\nabla_{\mathbf x} o \right\| = s(U; {\mathbf w}) \left| \hat{\delta}\right|.
\end{equation}
If the sensitivity $s(U; {\mathbf w})=\| \nabla_{\mathbf x} o \|$ is 1.0, then
\begin{equation}
\left\|\delta {\mathbf w} \right\| = \left| \hat{\delta}\right|,
\end{equation}
and the error signal is propagated efficiently without being shrunk or expanded through the neuron.

\begin{figure}[hbt]
\centerline{\includegraphics[scale=0.8]{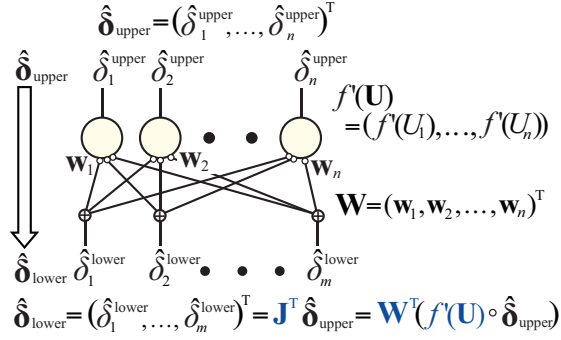}}
\caption{Backward (error signal) computation through a layer of neurons.}
\label{fig:layer_backward}
\end{figure}

From the view of group or layer of neurons as can be seen in Fig.~\ref{fig:layer_backward},
the relation between the error signal before and after the layer ${\hat{\boldsymbol \delta}}_{upper}$,
${\hat{\boldsymbol \delta}}_{lower}$ can be written as
\begin{equation}
{\hat{\boldsymbol \delta}}_{lower} = {\mathbf W}^\mathrm{T} \left(f'({\mathbf U})
                                   \circ {\hat{\boldsymbol \delta}}_{upper}\right).
\label{Eq:delta_hat}
\end{equation}
The Jacobian matrix as presented in Eq.~(\ref{Eq:JacobianMatrix}) can be modified as
\begin{eqnarray}
 {\mathbf J}({\mathbf U}; {\mathbf W}) &=&
 \begin{pmatrix}
 f'(U_1)w_{11} & \cdots & f'(U_1)w_{1m} \\
 \vdots & \ddots & \vdots \\
 f'(U_n)w_{n1} & \cdots & f'(U_n)w_{nm}
\end{pmatrix}
\\ \nonumber
 & = & \left(f'(U_1){\mathbf w}_1, \ldots , f'(U_n) {\mathbf w}_n\right)^\mathrm{T}.
\label{Eq:JacobianMatrix_Mod}
\end{eqnarray}
Then, Eq.~(\ref{Eq:delta_hat}) is changed to
\begin{equation}
{\hat{\boldsymbol \delta}}_{lower} = {\mathbf J}^\mathrm{T}({\mathbf U}; {\mathbf W}) {\hat{\boldsymbol \delta}}_{upper}.
\end{equation}
Accordingly, though the Jacobian is transposed,
the similar discussion as the relation between $\|d{\mathbf o}\|$ and $\|d{\mathbf x}\|$ in the forward computation
can be made.
The same assumptions are made for ${\mathbf W}$ and $f'(U)$ as before, and
${\hat{\boldsymbol \delta}}_{upper}$ is assumed to be a random vector
whose elements are i.i.d.~with mean 0.
The relation between the mean squared error signal vectors before and after the layer can be found as
\begin{equation}
\frac{1}{m}\mathit{E}\left [\left\| \hat{\boldsymbol \delta}_{lower} \right\|^2 \right ] 
= \frac{\langle s^2 \rangle}{n}\mathit{E}\left [\left\| \hat{\boldsymbol \delta}_{upper} \right\|^2 \right ],
\end{equation}
in the same way as for Eq.~(\ref{Eq:layer_do_dx}).
If the sensitivity of each neuron is 1.0,
\begin{equation}
\frac{\mathrm{RMS}\left[\left\|{\hat{\boldsymbol \delta}}_{lower}\right\| \right]}{\sqrt{m}}
= \frac{\mathrm{RMS}\left[\left\| {\hat{\boldsymbol \delta}}_{upper}\right\| \right]}{\sqrt{n}},
\end{equation}
and the error signals in backward computation do not either disappear or explode.
If the numbers of inputs and outputs are the same, RMS remains unchanged
before and after the layer as stated below:
\begin{equation}
\mathrm{RMS}\left[\left\|\hat{\boldsymbol \delta}_{lower}\right\|\right]
= \mathrm{RMS}\left[\left\|{\hat{\boldsymbol \delta}}_{upper}\right\|\right].
\end{equation}
Therefore, if the sensitivities of all network neurons can be controlled around 1.0,
it is expected that the error signals propagate backward without being shrunk or expanded.

\subsection{Learning Method (Sensitivity Adjustment Learning (SAL))}
In sensitivity adjustment learning (SAL), which we also propose in this paper,
each neuron has a small weight vector initially and updates them
to increase its sensitivity by hill-climbing that is the steepest ascent according to
\begin{equation}
\Delta {\mathbf w} = \eta_{SAL} \nabla_{\mathbf w} s(U;{\mathbf{w}}) = \eta_{SAL} \nabla_{\mathbf w} \left\| \nabla_{\mathbf x} o\right\|
\label{eq:hill_climbing}
\end{equation}
where $\eta_{SAL}$ is the learning rate for SAL.
From Eq.~(\ref{Eq:sensitivity}),
\begin{eqnarray}
\nabla_{\mathbf w} \| \nabla_{\mathbf x} o\| & = & \nabla_{\mathbf w} \left\{ f'(U) \left\| {\mathbf w}\right\| \right\}
\nonumber \\
& = & f'(U)\nabla_{\mathbf w} \left\| {\mathbf w}\right\| + \left\| {\mathbf w}\right\|\nabla_{\mathbf w} f'(U)
\nonumber \\
& = & f'(U)\frac{\mathbf w}{\left\| {\mathbf w} \right\|} + \left\| {\mathbf w} \right\| \nabla_{\mathbf w} f'(U)
\label{eq:grad_o}
\end{eqnarray}
Then the update rule can be written as
\begin{equation}
\Delta {\mathbf w} = \eta_{SAL} \left( f'(U)\frac{\mathbf w}{\left\| {\mathbf w} \right\|}
                         + \left\| {\mathbf w} \right\| \nabla_{\mathbf w} f'(U) \right).
\end{equation}
In the following simulations, we use $\tanh$ as activation function $f(\cdot)$ for all the neurons.
In this case, 
\begin{equation}
f'(U) = \frac{1}{\cosh^2(U)} = 1- o^2
\end{equation}
and so
\begin{eqnarray}
\nabla_{\mathbf w} f'(U) & = & \nabla_{\mathbf w} (1-o^2) \nonumber \\
& = & -2o f'(U)\nabla_{\mathbf w} U \nonumber \\
& = & -2o(1-o^2){\mathbf x}.
\end{eqnarray}
Then the update rule is as
\begin{equation}
\Delta {\mathbf w} = \eta_{SAL} (1-o^2) \left( \frac{\mathbf w}{\left\| {\mathbf w}\right\|} - 2o\left\|{\mathbf w}\right\|{\mathbf x} \right).
\label{Eq:Delta_w}
\end{equation}

In this equation, the term $\eta_{SAL} (1-o^2){\mathbf w}/\left\|{\mathbf w}\right\|$ is originated from
the first term of the right-hand side of Eq.~(\ref{eq:grad_o}), and is called `linear term'.
${\mathbf w}/\|{\mathbf w}\|$ is the unit vector whose direction is the same as ${\mathbf w}$.
Therefore, the sensitivity is increased directly by making the size of the weight vector greater
while keeping its direction.
Accordingly, if the neuron output $o$ is around the linear region, in which the output is around 0.0,
the weight vector ${\mathbf w}$ becomes greater at a constant rate decided by $\eta_{SAL}$.
The second term $-2 \eta_{SAL} (1-o^2) o\left\|{\mathbf w}\right\|{\mathbf x}$ is originated from the second term
of the right-hand side of  Eq.~(\ref{eq:grad_o}), and is called `non-linear term'.
This means the sensitivity is increased indirectly by making the magnitude of the internal state's
smaller for larger $f'(U)$.
This term works only after the neuron goes out of the linear region of the activation function.

Different from the case of weight, bias $\theta$ cannot increase the sensitivity directly but
can increase it indirectly by updating the bias so that the value $U$ becomes closer to 0.0.
Accordingly, the update rule for the bias is indicated as
\begin{equation}
\hspace{-7mm}
\Delta \theta = \eta_{SAL} \left\| {\mathbf w} \right\| \frac{\partial f'(U)}{\partial \theta}.
\label{eq:Delta_theta}
\end{equation}
Furthermore, assuming the activation function is $\tanh$, it can be rewritten as
\begin{equation}
\Delta \theta = -2\eta_{SAL} o (1-o^2) \left\|{\mathbf w}\right\|.
\end{equation}

In this paper, to control the sensitivity around 1.0,
SAL is applied only when the sensitivity is not greater than 1.0 in each neuron.
However, in practice, since the sensitivity is a function of $U$ (internal state after bias is added)
or output $o$, it fluctuates largely as $U$ or $o$ fluctuates.
Therefore, in this paper, the moving average of sensitivity $\overline {s}$ in each neuron is computed as
\begin{equation}
\overline{s}_n \leftarrow \beta \overline{s}_{n-1} + (1-\beta) s_n
\label{Eq:decay}
\end{equation}
where $\beta$ is a decay rate.
$n$ indicates the number of forward computations in the whole learning for each neuron,
and the initial value is set as $\bar{s}_{0}=0$ here.
In the case of an RNN, one neuron executes its forward computation repeatedly,
and so $n$ indicates the time step, but is not reset for a new pattern or epoch.
It is used as the criterion to decide the application of SAL in each neuron.

When supervised learning  is performed together with SAL learning,
error backpropagation (BP) learning based on stochastic gradient descent (SGD)
including BPTT is used such as
\begin{equation}
\Delta {\mathbf{w}} = -\eta_{BP} \nabla_{\mathbf{w}} E
\label{Eq:Dw_BP}
\end{equation}
where $E$ is the squared error when a training signal vector ${\mathbf d}$
is given and is expressed as
\begin{equation}
E = \frac{1}{2} ({\mathbf d} - {\mathbf o}^{out})^2
\end{equation}
where ${\mathbf o}^{out}$ is the output vector in the output layer.
In this paper, assuming simple on-line learning, no batch learning is done,
and the network is trained for each presented pattern one by one.
However, we expect no hurdles for the extension to batch learning
because of the SAL's local learning nature.

\begin{wrapfigure}[31]{r}[0mm]{85mm}
  \centering
  \vspace{-3mm}
  \includegraphics[keepaspectratio,width=50mm]{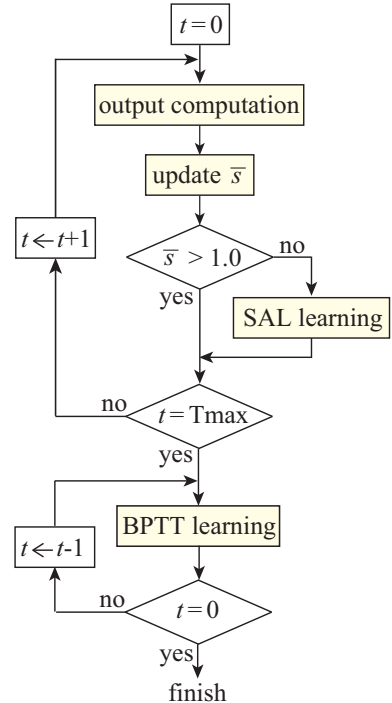}
  \captionsetup{width=72mm}
  \caption{Flow chart for the parallel learning of SAL and BPTT for one pattern presentation.
SAL is applied only when the moving average of the sensitivity $\overline s$ is not greater than 1.0.
$T_{max}$ is the timing of output.
This conditional branch and SAL itself are performed individually in each neuron.}
\label{fig:flow_chart}
\end{wrapfigure}
Since the sensitivity fluctuates largely as mentioned, the satisfaction of $\bar{s}=1$ in each neuron
does not fully guarantee that the error signals will not explode in the backward computation.
Then, we introduce an additional technique to avoid the exploding gradient problem here.
The function $\tanh$ is equivalent to the identity function when the input is around 0.0,
but the output is saturated when the absolute value of the input is large.
Then, in the backward computation of the error signals, 
the function $\tanh$ is applied after multiplicting the derivative $f'(U)$.
Instead of applying Eq.~(\ref{Eq:multi_derivative}), the error signal is computed as
\begin{equation}
\delta = \tanh(\hat{\delta} f'(U)),
\label{Eq:tanhinBP}
\end{equation}
and the weight vector is updated as
\begin{equation}
\Delta {\bf w} = \eta_{BP} \delta {\bf x}.
\end{equation}

In this paper, when we apply SAL with BP or BPTT,
the weights are updated by SAL just after the forward computation at each timing in each neuron
when the moving average of its sensitivity  $\overline s$ is not greater than 1.0.
Once the network finishes all the forward computations and SAL learning for the presented pattern,
BP or BPTT is applied in the backward computation.
Although the error signal is not propagated backward using the original weights but using those after the SAL learning,
we think the way is more effective than applying either SAL or (BP or BPTT) for one forward computation.
The flowchart is shown in Fig.~\ref{fig:flow_chart} for the case of learning of an RNN with BPTT
as an example.
\newpage

%% file: Simulations.tex
\section{Simulations}
The following simulations are roughly divided into two parts.
In the first part, as described in Section~\ref{sec:sim_chaos},
only sensitivity adjustment learning (SAL) is applied to recurrent neural networks (RNNs).
Then, generation and control of chaotic dynamics are examined,
and the relation between sensitivities and maximum Lyapunov exponent is focused on.
In the second part described in Section~\ref{sec:sim_BP}, 
SAL is applied with supervised learning using BP or BPTT.
Then the performance is observed, and the learning process is analyzed.
In all the simulations, hyperbolic tangent is used as the activation function of each neuron.

\subsection{Generation and Control of Chaos}\label{sec:sim_chaos}
At first, we show the generation of chaos dynamics in RNNs and adjustment
of its chaoticity by sensitivity adjustment learning (SAL)
for various network architectures.

\begin{table}[b]
  \begin{center}
    \caption{Parameters for chaos generation by SAL using a flat RNN}
    \vspace{3mm}
      \begin{tabular}{c||c} \hline
        Number of neurons & 100 or 30\\ \hline
        Connection rate (\%) & 100 or 30\\ \hline
        Initial connection weights	& 
       \begin{tabular}{c}Uniformly random\\$[-0.01, 0.01]$ \end{tabular}
        \\ \hline
        Learning rate $\eta_{SAL}$ in Eq.~(\ref{Eq:Delta_w}) & 0.00002\\ \hline
	 Decay rate $\beta$ in Eq.~(\ref{Eq:decay}) & $ 0.99 $\\ \hline
	 Perturbation (interval, size) & $ (1000, 0.001)$\\ \hline
      \end{tabular}
    \label{table:param_flatRNN}
  \end{center}
\end{table}
Before entering the simulation result, the way to estimate
the maximum Lyapunov exponent is explained~[\cite{Sprott}].
At every 100 steps ($t=0,100,200....$), two internal states, ${\mathbf u}^{(1)}_0={\mathbf u}_t$ and
${\mathbf u}^{(2)}_0={\mathbf u}_t+{\mathbf{rnd}}$ where ${\mathbf{rnd}}$ is a small random vector
whose Euclidian norm is $10^{-3}$,
are prepared.
The two states are updated separately without updating the weights.
At each step $\tau$, the distance ratio between before and after one step update is calculated as
\begin{equation}
l_\tau = ln\left(\frac{{\mathbf u}^{(2)}_\tau-{\mathbf u}^{(1)}_\tau}{10^{-3}}\right).
\end{equation}
After that, the distance is normalized to $10^{-3}$ as
\begin{equation}
{\mathbf u}^{(2)}_\tau \leftarrow \frac{10^{-3}}{\left\|{\mathbf u}^{(2)}_\tau-{\mathbf u}^{(1)}_\tau\right\|}
    \left({\mathbf u}^{(2)}_\tau-{\mathbf u}^{(1)}_\tau\right)+{\mathbf u}^{(1)}_\tau.
\end{equation}
Finally, the maximum Lyapunov exponent is computed using the 1,000 steps
of the forward computation after 100 steps to eliminate the influence of the initial perturbation as
\begin{equation}
\lambda = \frac{1}{1,000}\sum_{\tau=101}^{1,100} l_\tau.
\end{equation}
When the network has two layers, that is computed at the 1st layer.

\subsubsection{Case of Flat RNN}
Firstly, investigation using a flat RNN with no layer structure is introduced.
Table~\ref{table:param_flatRNN} presents the parameters.
The initial connection weights are small and uniform random numbers, and no bias is used here.
The learning rate is small to see the dynamics for each stage during learning.
A small random perturbation is added to the internal state vector ${\mathbf u}$ at every 1000 steps
from the 1st step as a trigger of activations.
\begin{figure}[t]
\centerline{\includegraphics[scale=0.37]{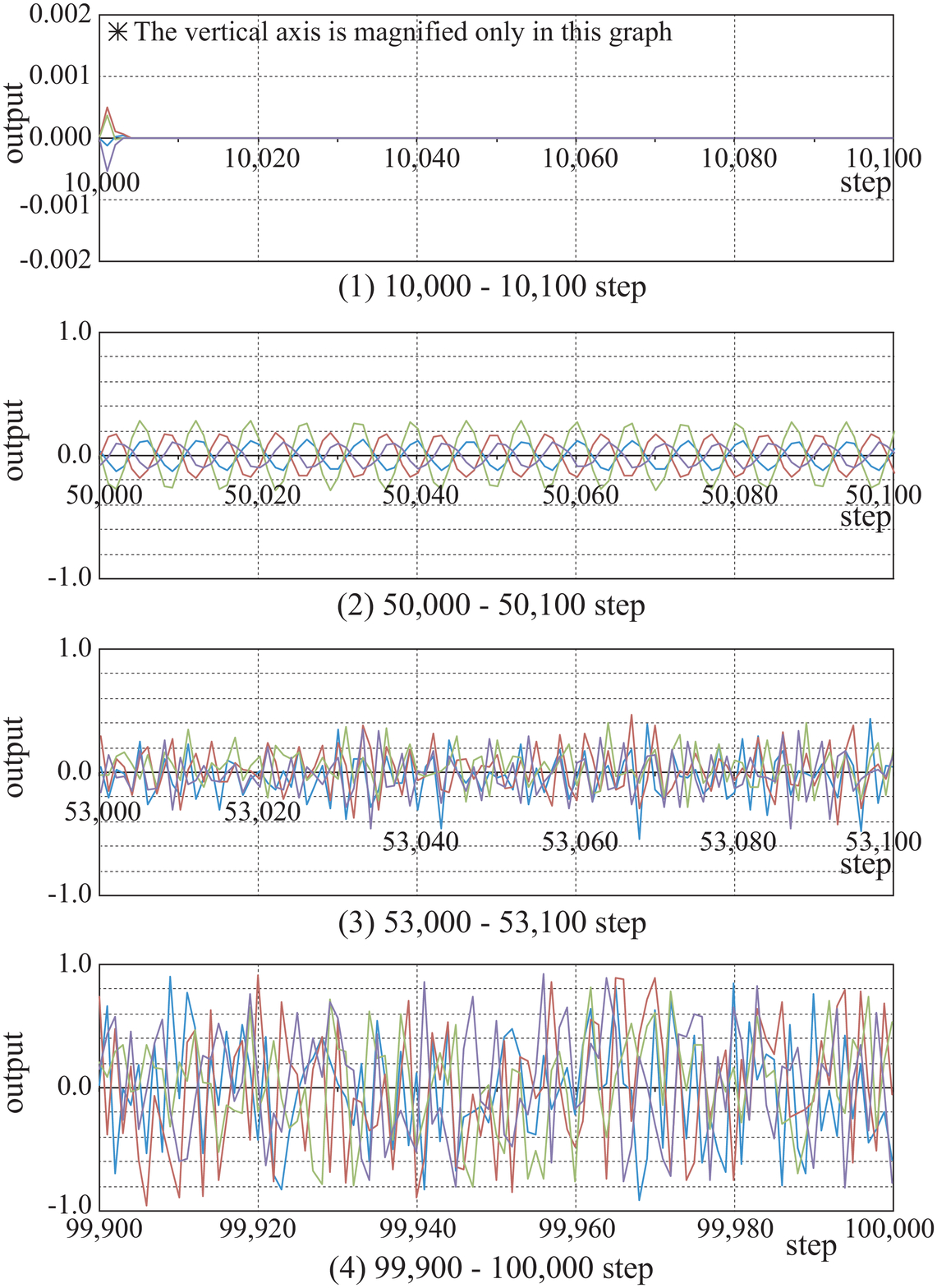}} 
\caption{Change of four sample time-series of four outputs as learning progresses from (1) to (4)
when applying SAL to a flat RNN.}
\label{fig:output}
\end{figure}

In the first simulation, using an RNN with fully connected 100 neurons,
the learning process is shown in detail.
Fig.~\ref{fig:output} shows the output change of four sample neurons at each of the four stages
from (1) to (4) during learning.
In early phase of learning, as shown in Fig.~\ref{fig:output} (1),
although a small perturbation is added at every 1,000 steps, the output is decayed soon.
Around the 50,000th step, the outputs change almost periodically
as shown in Fig.~\ref{fig:output} (2),
and around the 53,000th step, the outputs change irregularly as shown in Fig.~\ref{fig:output} (3).
Around the 100,000th step, as shown in Fig.~\ref{fig:output} (4),
the outputs still change irregularly, but the value range is greater than in (3).

\begin{figure}[thbp]
\centerline{\includegraphics[scale=0.44]{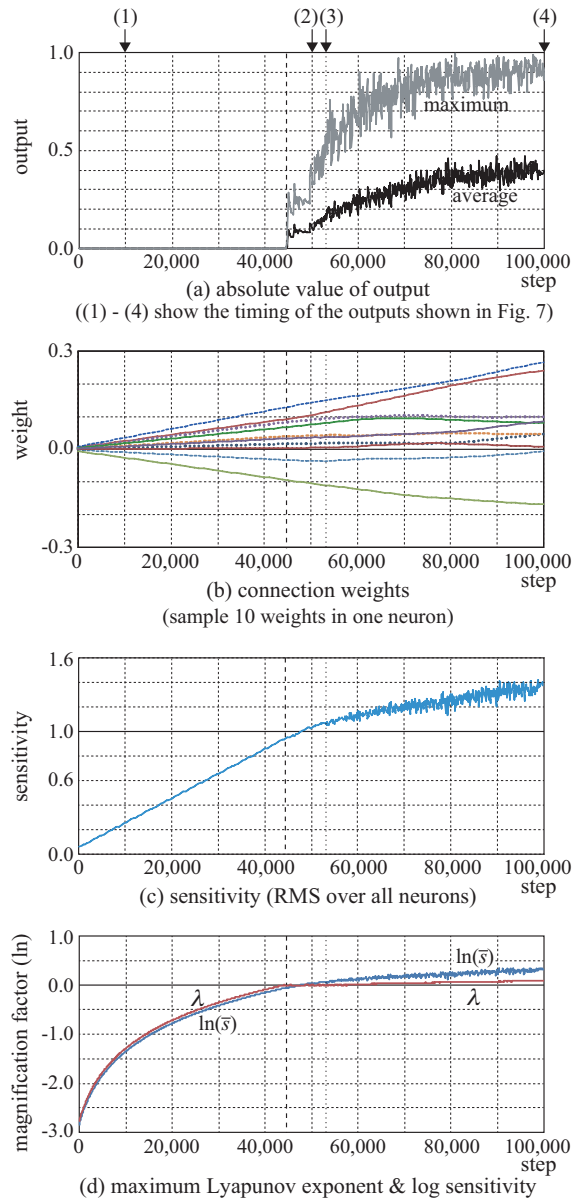}}
\vspace{3mm}
\caption{Learning process when applying SAL to a flat RNN.
(a) The maximum and average absolute value over all the outputs. (b) Sample connection weights.
(c) RMS of sensitivities over all the neurons. (d) Maximum Lyapunov exponent and log sensitivity
(logarithm of the value in (c)). All the data are plotted at every 100 steps.
The thick vertical broken line shows the boundary of whether the outputs decay to 0.0 or not.}
\label{fig:learning}
\end{figure}
Fig.~\ref{fig:learning} shows how the learning progressed from various aspects.
Fig.~\ref{fig:learning}(a) provides the maximum and mean absolute value of the output
over all the 100 neurons.
Before around the 45,000th step indicated by the thick vertical broken line,
all the outputs were almost 0.0 though small perturbations were added.
Around the 45,000th step, the outputs increased suddenly.
After that, they increased gradually with some fluctuations,
and finally they varied in most of the value range as can be seen in Fig.~\ref{fig:output}(4).
Fig.~\ref{fig:learning}(b) shows how some sample weights changed during learning.
Until around the 45,000th step when the output of all the neurons is in the linear region,
the absolute value of each weight grows constantly,
which is consistent with Eq.~(\ref{Eq:Delta_w}).
After that, the absolute weight value still tends to increase, but sometimes it decreases,
and sometimes the magnitude relation between two weights is switched
by the influence of non-linear term in Eq.~(\ref{Eq:Delta_w}).

\begin{figure}[tb]
\centerline{\includegraphics[scale=0.37]{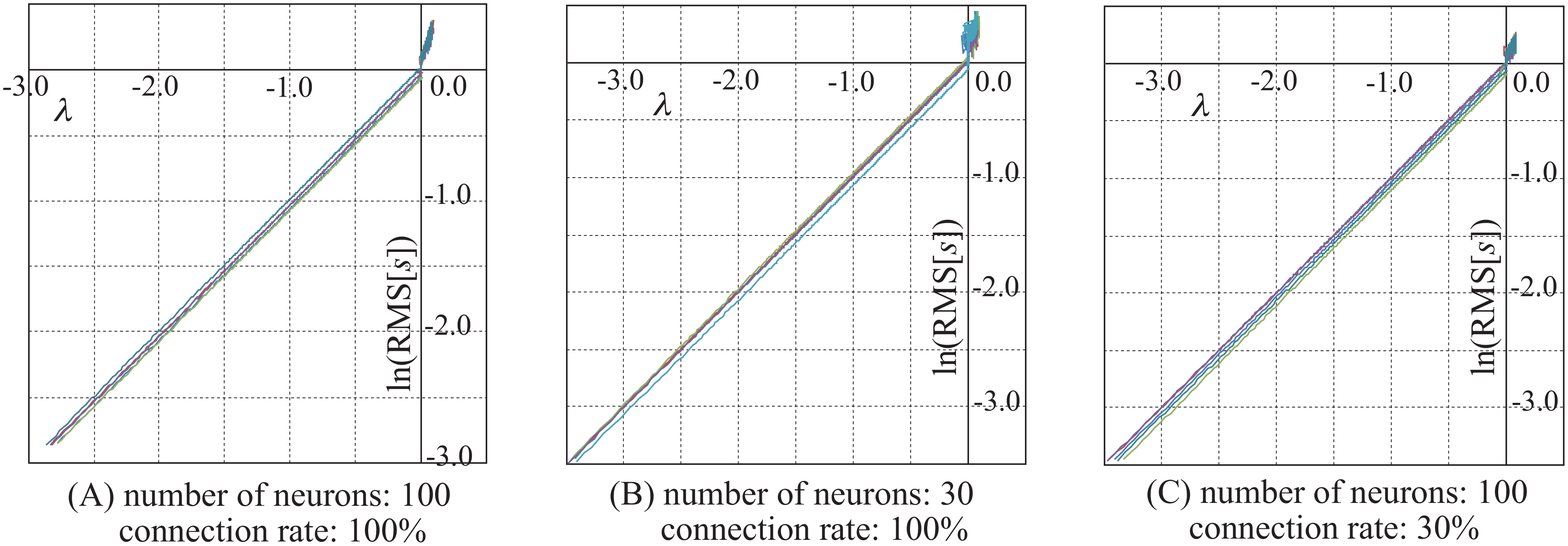}} 
\caption{Relation between the maximum Lyapunov exponent and log-sensitivity during SAL for 5 cases,
varying (B) number of neurons or (C) connection rate.}
\label{fig:L-S-1L}
\end{figure}
Fig.~\ref{fig:learning}(c) presents the change of RMS of the sensitivities over all the neurons
during learning.
It increases almost linearly until it reaches 1.0 right after the neurons take non-zero values.
After that, the increase rate becomes smaller, and the value is fluctuating.
To compare it with the maximum Lyapunov exponent,
which shows the chaoticity of the network dynamics,
the RMS sensitivity is plotted on the log-scale in Fig.~\ref{fig:learning}(d).
We call the value `log-sensitivity' here.
The maximum Lyapunov exponent $\lambda$ is also plotted on the same graph.
They are almost the same until around the 45,000th step.
Afterward, the slope for $\lambda$ becomes less than that for log-sensitivity,
but $\lambda$ is still increasing.
Then Fig.~\ref{fig:L-S-1L}(A) shows the relationship between the maximum Lyapunov exponent $\lambda$
and log-sensitivity $ln(RMS[s])$ by plotting them on the $x$- and $y$-axis respectively.
Five lines indicate five cases with different initial connection weights decided randomly.
The relations are similar in all the five cases, and
when the log-sensitivity reaches 0.0, the maximum Lyapunov exponent reaches almost 0.0.

Then the number of neurons or connection rate is varied and it is examined
whether the same relationship can be seen between the maximum Lyapunov exponent
and the log-sensitivity.
Fig.~\ref{fig:L-S-1L}(B) shows the results when decreasing the number of neurons to 30,
and Fig.~\ref{fig:L-S-1L}(C) shows the results when decreasing the connection rate to 30\%.
In both figures, the maximum Lyapunov exponent is almost identical to the log-sensitivity
until 0.0 as in the fully connected RNN with 100 neurons.
Only the difference is that the initial value is almost $-3.5$ in (B) and (C), but originally
it is around $-2.8$ in (A).
Those are almost the same as the expected value $ln \left(\mathrm{RMS}\left[ \| {\mathbf w} \| \right]\right)$
derived from Eq.~(\ref{Eq:sensitivity}).
Since the initial weight values are decided as a uniform random number from $-0.01$ to $0.01$,
$\mathrm{Var}\left[w_i\right]=10^{-4}/3$.
Expected number of inputs is 100 in (A), 30 in (B) and (C), and $f'(U)\approx 1.0$.

\vspace{2mm}
\subsubsection{Case of Two-Layer RNN}
\begin{figure}[h]
\vspace{3mm}
\centerline{\includegraphics[scale=0.5]{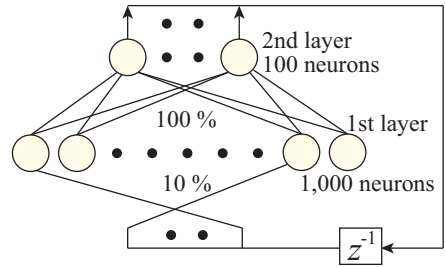}} 
\caption{Two-layer RNN used in this paper. SAL was applied to all the neurons in the network.
\vspace{-5mm}}
\label{fig:2L-RNN}
\end{figure}

The relationship between the maximum Lyapunov exponent and
log-sensitivity is examined in a multi-layer recurrent neural network (RNN).
Here, a two-layer RNN as shown in Fig.~\ref{fig:2L-RNN} is employed.
The first layer has 1000 neurons, and the second layer has 100 neurons.
The connection rate from the first layer to the second layer is 100\%.
In contrast, the feedback connection rate from the second layer to the first layer is 10\%.
The expected number of connections is 10 for the first layer neurons,
while 1000 for the second layer neurons.
No bias is used.
Table~\ref{table:param_2L-RNN} summarizes the parameters.
\begin{table}[b]
  \begin{center}
    \caption{Parameters for the chaos generation by SAL using a 2-layer RNN}
    \vspace{3mm}
      \begin{tabular}{c||c} \hline
        Number of neurons & 1000 (1st) 100 (2nd)\\ \hline
        Connection rate (\%) & 
        \begin{tabular}{c}100 (2nd $\leftarrow$ 1st)\\
        10 (1st $\leftarrow$ 2nd) \end{tabular} \\ \hline
        Initial connection weights	& 
        \begin{tabular}{c}
            Uniformly random\\
            $[-0.03, 0.03]$
        \end{tabular}
        \\ \hline
        Learning rate $\eta_{SAL}$ in Eq.~(\ref{Eq:Delta_w}) & 0.00002\\ \hline
	 Decay rate $\beta$ in Eq.~(\ref{Eq:decay}) & $ 0.99 $\\ \hline
	 Perturbation (interval, size) & $ (1000, 0.001)$\\ \hline
      \end{tabular}
    \label{table:param_2L-RNN}
  \end{center}
\end{table}

\begin{figure}[tbp]
\begin{center}
\vspace{7mm}
\includegraphics[scale=0.37]{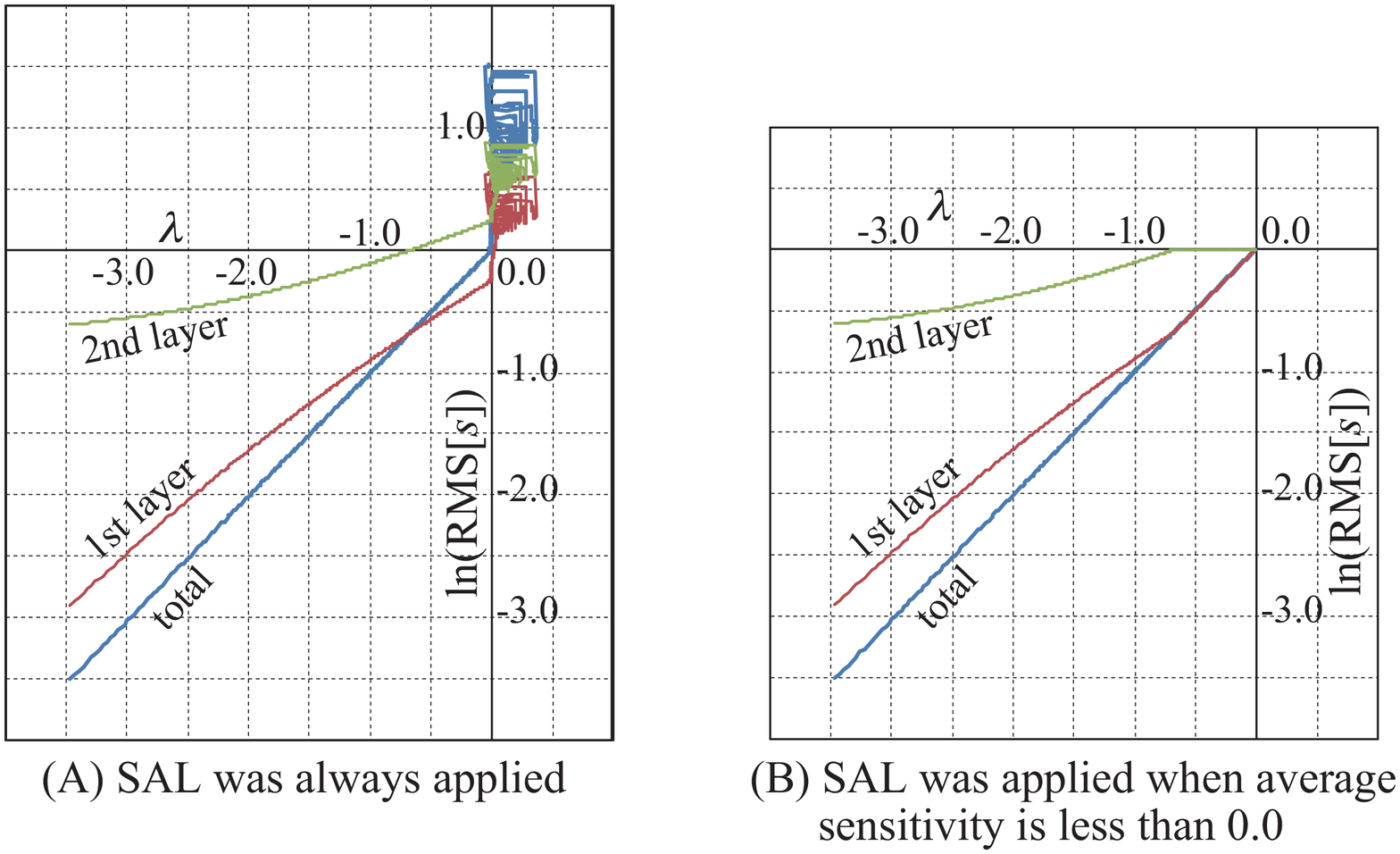}
\end{center}
\caption{Relation between the maximum Lyapunov exponent and log-sensitivity during SAL.
  (1000--100 neurons, 10\%--100\% connection rate)
  (B) shows the relation when the target sensitivity was set to 1.0.
}
\label{fig:L-S-2L}
\end{figure}

Fig.~\ref{fig:L-S-2L}(A) shows the relationship between each layer's log-sensitivity
and the maximum Lyapunov exponent.
Here `total log-sensitivity' is introduced that is the sum of the log-sensitivities of both layers.
The sensitivity value is different between the two layers
because the number of connections is different largely.
As learning progresses, the log-sensitivities increased,
and in the second layer, it became greater than 0.0
though the maximum Lyapunov exponent was still negative.
However, it can be seen that the total log-sensitivity is almost the same
as the maximum Lyapunov exponent, and when the total log-sensitivity
reached 0.0, the maximum Lyapunov exponent reached around 0.0.

Then to adjust the network dynamics to be around the edge of chaos,
SAL was stopped in each neuron when the moving average of its sensitivity
in Eq.~(\ref{Eq:decay}) reached 1.0.
Fig.~\ref{fig:L-S-2L} (B) displays the results.
It shows that the second layer's log-sensitivity did not become greater than 0.0
by stopping SAL in each of the 2nd layer neurons.
Finally the network dynamics reached around the edge of chaos,
which is the dynamical state with the maximum Lyapunov Exponent $\lambda = 0.0$,
and maintained it by stopping SAL also in each of the 1st layer neurons.

\subsection{Solving the Vanishing Gradient Problem}\label{sec:sim_BP}
Next, let us focus on how sensitivity adjustment learning (SAL) works to avoid
vanishing gradient problems in the backward computation for gradient-based learning.
Here, SAL is applied in each neuron only when the sensitivity is not greater than 1.0,
following the flowchart as presented in Fig.~\ref{fig:flow_chart}.

In the following, two cases of supervised learning are shown.
The first one is a recurrent neural network (RNN) solving a problem with long-term dependency,
and the network is trained by BPTT (error Back Propagation Through Learning).
The second one is a deep feedforward network (DFNN) trained by BP (error Back Propagation).
In each case, in order to see how SAL works in gradient-based learning,
a simple learning problem and stochastic gradient descent (SGD) is used.
Therefore, learning is not applied to a batch or mini-batch but applied for each pattern presentation.

\subsubsection{Case of RNN Solving a Long Time-Lag Problem}
\begin{figure}[t]
\centerline{\includegraphics[scale=0.85]{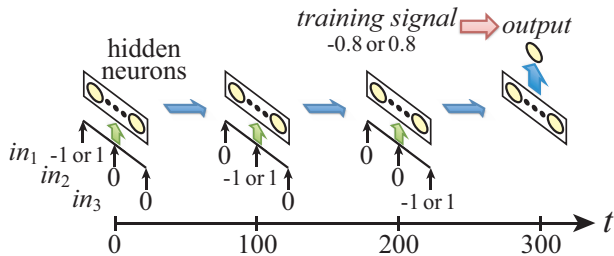}}
\caption{Sequential 3-bit parity problem with a lag of 300 steps.
Three inputs were given in turn with intervals of 100 steps.}
\label{fig:BPTT}
\end{figure}

Here, to see how SAL affects gradient-based supervised learning,
a simple 3-layer Elman-type RNN is used whose hidden outputs are fed back to themselves
at the next time step.
As a simple learning problem, a sequential 3-bit parity problem with a lag of 300 steps
as shown in Fig.~\ref{fig:BPTT} is given.
In the problem, three inputs are given sequentially at every 100 steps,
and the training signal is given 300 steps after the timing of the first input.
Eight patterns are presented in one epoch.
\begin{table}[t]
  \begin{center}
    \caption{Parameters for supervised learning using SAL and BPTT in an RNN.}
    \vspace{3mm}
      \begin{tabular}{c|c||c} \hline
        \multicolumn{2}{c||}{Number of neurons (input, hidden, output)}&(3, 20, 1)\\ \hline
        \renewcommand{\arraystretch}{0.9}
        \multirow{3}{*}{\begin{tabular}{c}Initial\\ connection weights\\ \small{(uniformly random)} \end{tabular}} 
        \renewcommand{\arraystretch}{1.0}
        & input $\rightarrow$ hidden & 0.0 \\
        & hidden $\rightarrow$ output & $[-0.3, 0.3]$ \\
        & hidden $\rightarrow$ hidden & $[-0.1, 0.1]$ \\ \hline
        \multicolumn{2}{c||}{Learning rate $\eta_{SAL}$ in Eq.~(\ref{Eq:Delta_w})}&0.0002\\ \hline
        \multirow{3}{*}{\begin{tabular}{c}Learning rate\\ $\eta_{BP}$ in Eq.~(\ref{Eq:Dw_BP}) \end{tabular}} 
        & input $\rightarrow$ hidden & 0.4 \\
        & hidden $\rightarrow$ output & 0.1 \\
        & hidden $\rightarrow$ hidden & 0.00004 \\ \hline
         \multicolumn{2}{c||}{Decay rate $\beta$ in Eq.~(\ref{Eq:decay})}&0.999\\ \hline
      \end{tabular}
    \label{table:parameters_RNN}
  \end{center}
\end{table}

Table~\ref{table:parameters_RNN} presents the parameters used here.
Here, a bias was used in each neuron.
Its initial value and learning rate were decided
in the same way as the connection weights in the same layer.
For the hidden neurons, they were decided in the same way as the feedback connection weights.
The propagated error signals and sensitivities during learning are observed
as well as the learning performance.
One hundred simulations are performed with different random initial connection weights.
Here, only the feedback weight vector ${\mathbf w}_{FB}$ is used to compute the sensitivities
according to Eq.~(\ref{Eq:sensitivity}), and they are computed at every timing except for $t=0$
in each hidden neuron.
The learning is considered successful if the absolute value of the error becomes less than
$0.01$ for each of the eight patterns within 1000 epochs.

\begin{figure}[t]
\centering
\vspace{4mm}
\centerline{\includegraphics[scale=0.42]{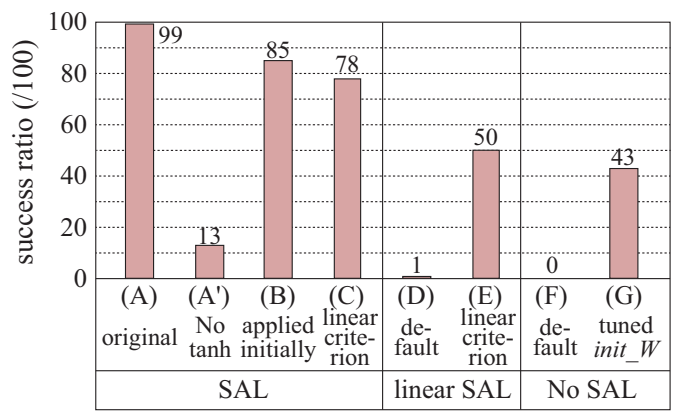}} 
\caption{Comparison of success ratio in 100 simulation runs among various conditions
in Table~\ref{table:SimulationList} in the supervised learning using SAL and BPTT in an RNN.
}
\label{fig:ComparisonRNN}
\end{figure}

\begin{table}[t]
  \begin{center}
    \caption{List of simulations performed for comparison
                 in the supervised learning of sequential 3bit-parity problem using SAL+BPTT in an RNN.}
    \vspace{3mm}
      \begin{tabular}{c||c|c|c} \hline
      Case & SAL & Criterion & Others\\ \hline \hline
      (A) & Nonlinear & Nonlinear & `original' \\ \hline
      (A') & Nonlinear & Nonlinear & No $\tanh$ in BPTT \\ \hline
      (B) & Nonlinear & Nonlinear & Applied only initially\\ \hline
      (C) & Nonlinear & Linear & \\ \hline
      (D) & Linear & Nonlinear&\\ \hline
      (E) & Linear & Linear&\\ \hline
      (F) & Not applied & -- &\\ \hline
      (G) & Not applied & -- & Tuned $init\_W$\\ \hline
       \end{tabular}
    \label{table:SimulationList}
  \end{center}
\end{table}
Here, to see the effect of SAL and the reason why the combination of SAL and BPTT works well,
the performance is compared among the various conditions
from the case (A) to (G) as shown in Table~\ref{table:SimulationList}.
Each of the conditions will be explained below.
Fig.~\ref{fig:ComparisonRNN} shows the success ratio for each case.

\begin{figure*}[tbhp]
\centering
\centerline{\includegraphics[scale=0.57]{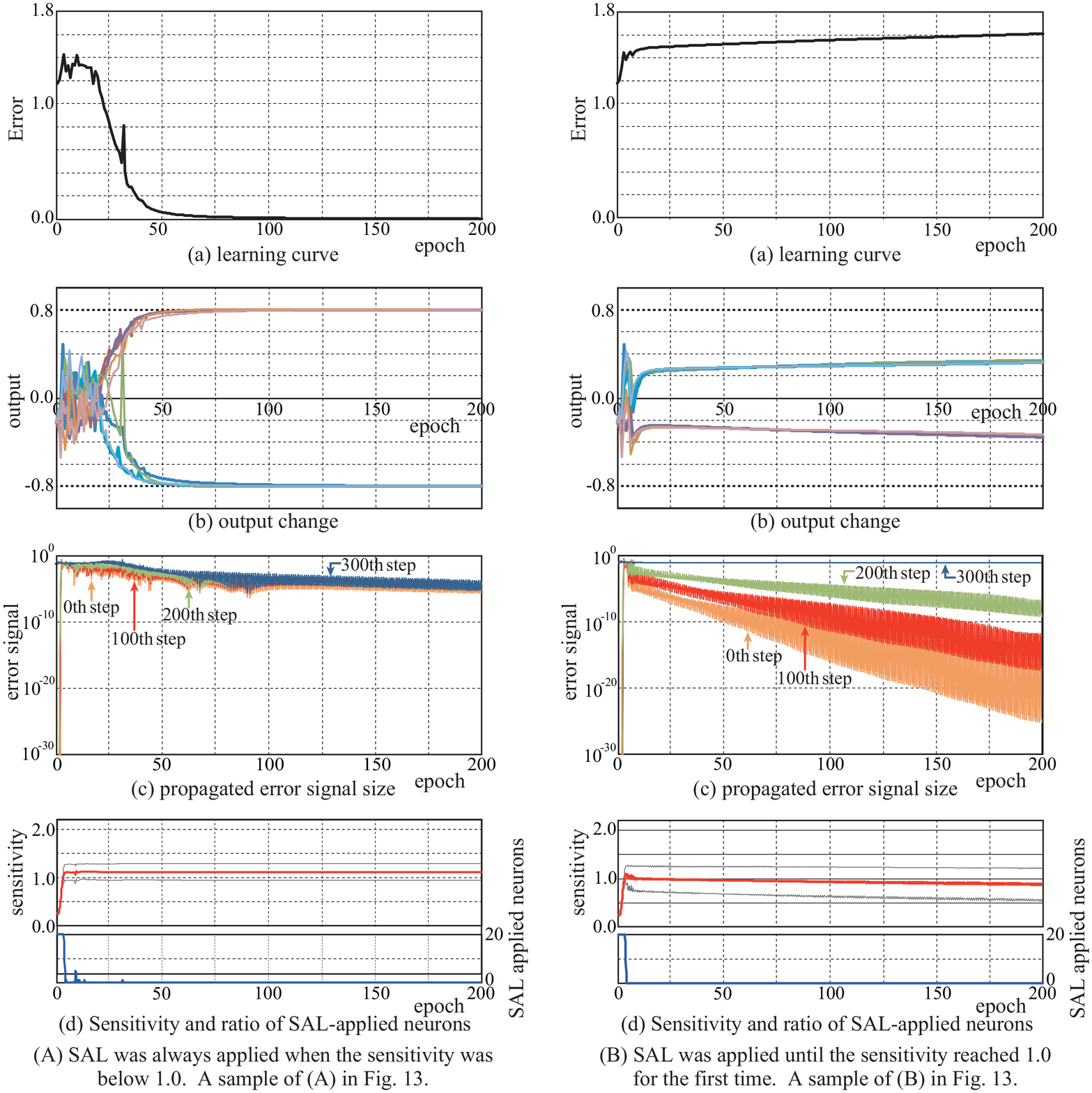}}
\vspace{2mm}
\caption{Sample learning process when SAL was applied in an RNN together with BPTT
and its comparison between two cases (A) and (B) in Table~\ref{table:SimulationList}.
In (B), SAL was not applied again in each neuron once its sensitivity reached 1.0.
The initial connection weights are the same between (A) and (B).
From the top, (a) change of RMS error over eight patterns,
(b) change in the network output for each of eight patterns,
(c) RMS of the magnitude of propagated error signal over all the neurons at several timings
in 300 steps of backward computation for BPTT.
(d) Upper: distribution of sensitivity over all the neurons and steps.
Three lines show the mean (center) and standard deviation.
Lower: number of neurons in which SAL was applied at least once.
(c) and (d) are plotted for each pattern presentation,
and so eight points are plotted in order for each epoch.
In the 0th epoch, no learning is applied.}
\label{fig:Learning_RNN}
\end{figure*}

First of all, let us see the case (A) where SAL and BPTT are applied normally,
and this case is called `original' for the following comparisons.
There is only one failure in total 100 runs,
but even in the failure run, the network could learn it in 3000 epochs.
Fig.~\ref{fig:Learning_RNN} (A) shows the learning process of a standard sample run
whose learning is the 46th fastest in the 100 runs.
(These initial weights will be used later as a sample of failures in other cases)
In Fig.~\ref{fig:Learning_RNN}, subfigures~(a) and (b) depict the learning curve
and the change in the network output for each of the eight patterns during learning.
Note that in the 0th epoch that includes eight pattern presentations,
no learning was applied to show the output and error signals before learning.
As shown in Fig.~\ref{fig:Learning_RNN}(A)-(a), after the 20th epoch, the error decreased gradually
except when it temporarily increased around the 30th epoch.

Subfigure~(c) in Fig.~\ref{fig:Learning_RNN} shows
how the RMS of the error signal $\delta$ changed in BPTT in the hidden layer during learning.
Before learning, the error signal at the 300th step, which is the output timing, was around $10^{-1}$,
but for the other input timings (200th, 100th, 0th step), it was far less than $10^{-30}$.
At the 0th step, it was less than $10^{-160}$ actually.
That indicates the gradient vanished through the backward propagation.
However, by applying SAL at every step, the error signals increased rapidly,
and in the 2nd epoch, they reached the same order as that at the output timing.
Then they stopped to increase.
After around 30 epochs of learning, all the four error signals gradually decreased
as the final error in (a) decreased.

Subfigure~(d) in Fig.~\ref{fig:Learning_RNN} shows the average
and standard deviation of sensitivities over all the 20 hidden neurons and all the 300 steps during learning.
The lower graph in the subfigure~(d) shows the number of neurons to which SAL was applied
even once during the 300 steps.
As shown in Fig.~\ref{fig:Learning_RNN}(A)-(d), soon after the learning began, the sensitivity increased 
and its average reached 1.0.
Then each neuron stopped SAL, and the value did not change so much,
keeping the average a bit greater than 1.0.
However, SAL was applied in some neurons afterward
when their average sensitivity decreased below 1.0.
It is also noticed that the sudden change in the output in Fig.~\ref{fig:Learning_RNN}(A)-(b)
mentioned above was caused by the SAL application around the 30th epoch.

On the other hand, in the case (F) when only performing BPTT without SAL,
no successful run can be seen as in Fig.~\ref{fig:ComparisonRNN}(F).
In RNNs such as a reservoir, the spectral radius of the feedback weight matrix
is usually tuned manually to control the network dynamics before learning. 
Then the spectral radius of ${\mathbf W}_{FB}$ was increased by 0.01 from 1.0 in the case (G).
The maximum success ratio was 43 when the spectral radius is 1.38 as in
Fig.~\ref{fig:ComparisonRNN}(G),
but the ratio was still far below the case when SAL was applied.

To analyze how SAL affects the learning positively,
five other cases from (A') to (E) in Table~\ref{table:SimulationList} were simulated.
Before entering the analysis of SAL itself, let us see the effect of applying $\tanh$ function
in the backward error computation as expressed in Eq.~(\ref{Eq:tanhinBP}).
Fig.~\ref{fig:ComparisonRNN}(A') shows the success ratio when SAL is applied
without using $\tanh$ function in error backpropagation in BPTT.
In the failure cases, although each neuron stopped to apply SAL
when its average sensitivity was greater than 1.0, the gradient or propagated error signals exploded.
The reason could be fluctuation of the sensitivity due to the term $f'(U)$ in Eq.~(\ref{Eq:sensitivity})
and/or the delay due to the average computation in Eq.~(\ref{Eq:decay}).
Therefore, to avoid such a gradient explosion, it is a good idea to use $\tanh$
also in the backward error computation.
In the following, $\tanh$ is always used in the backward computation.

At first, the effect of continuous learning is shown.
In the case of (B), SAL was applied only until the sensitivity reached 1.0 for the first time.
As shown in Fig.~\ref{fig:ComparisonRNN}(B), the ratio was worse than the `original' case (A).
Fig.~\ref{fig:Learning_RNN} (B) shows the learning process
when the initial weights were the same as the `original' case (A), but the learning failed
as can be seen in Fig.~\ref{fig:Learning_RNN}~(B)-(a,b).
As shown in Fig.~\ref{fig:Learning_RNN}(B)-(c), 
the propagated error signals reached $10^{-1}$ order at the second epoch as well as the case of (A).
However, soon after that, the error signal at the earlier steps decreased more
even though the error signal at the output timing (300th step) was not decreased.
As shown in Fig.~\ref{fig:Learning_RNN}{(B)-}(d), the mean sensitivities
became more than 1.0 once, but they decreased below 1.0 after that.
By comparing with the `original' case of (A), it is suggested that
SAL is useful not only at the beginning of learning but also works
to prevent the loss in sensitivities caused by BPTT during learning.

Secondly, the effect of nonlinear part $f'(U)$ of the sensitivity expressed in Eq.~(\ref{Eq:sensitivity})
is shown.
The nonlinear property influences mainly two parts in SAL.
One of them is the second term in the second parenthesis in Eq.~(\ref{Eq:Delta_w}) for weight update
in SAL itself.
The other part is that the sensitivity including $f'(U)$ is used as the criteria to decide
whether SAL is applied or not.
Here, the results of four combinations of the two conditions, each of which is linear or nonlinear, are shown.
The case when the non-linear term in Eq.~(\ref{Eq:Delta_w}) is deleted in SAL is called `linear SAL'. 
The case when $|{\mathbf w}_{FB}|$ is used as the criterion to apply SAL
instead of using sensitivity $f'(U)|{\mathbf w}_{FB}|$, is called `linear criterion'.
As can be seen in Table~\ref{table:SimulationList}, both are nonlinear in the `original' case (A).
In the case of (C), SAL itself is nonlinear, but criterion is linear.
Then SAL was stopped when $|{\mathbf w}_{FB}|$ was greater than 1.0.
The success ratio for the case (C) is less than the `original' case (A)
as shown in Fig.~\ref{fig:ComparisonRNN}.
Since $f'(U)$ is usually less than 1.0, the adjustment target value should be greater than 1.0
to keep the sensitivity to be around 1.0, but it is not easy to find the optimal target value. 

\begin{figure*}[t]
\centering
\centerline{\includegraphics[scale=0.57]{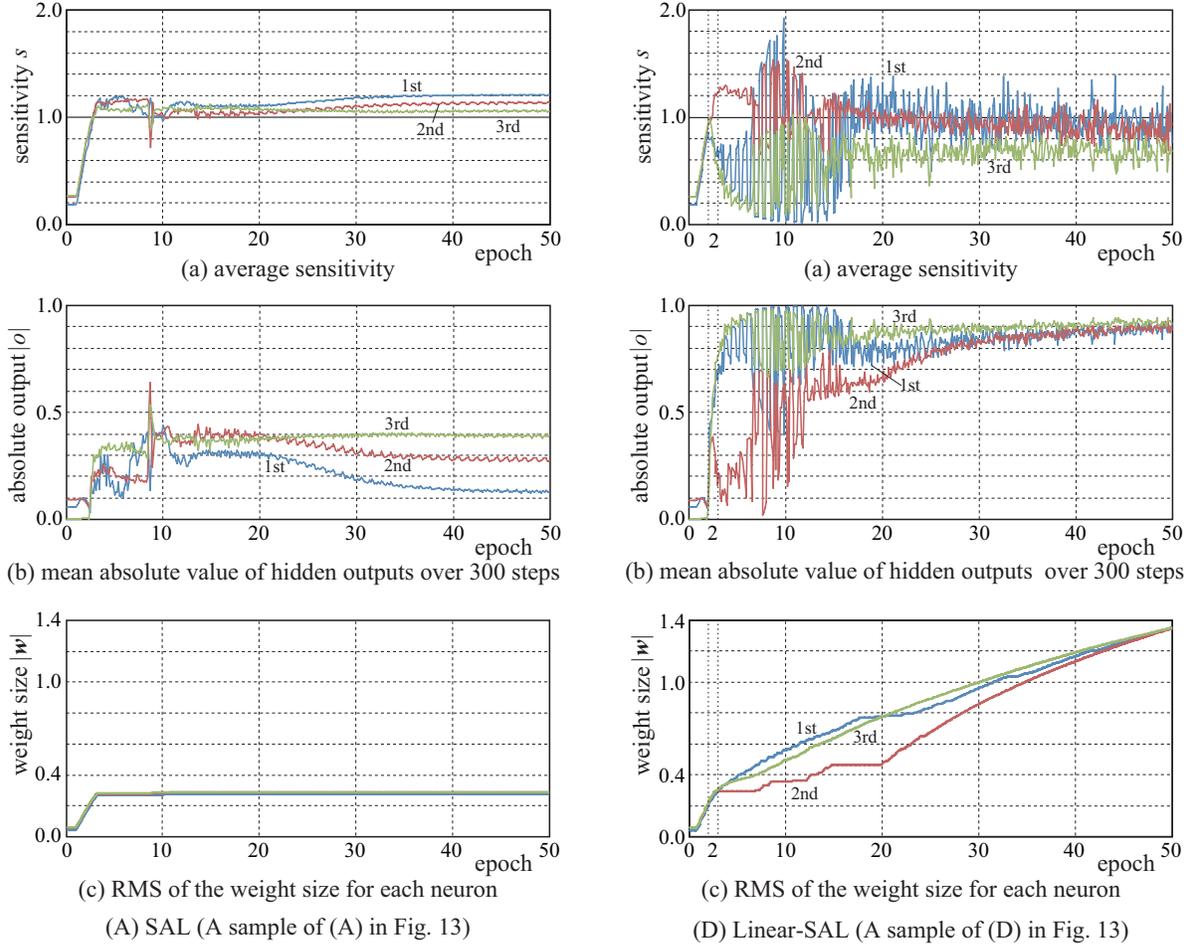}} 
\caption{Comparison of the early stage of the learning process (sensitivity $s$, absolute output $|o|$,
weight size $|{\mathbf w}|$ of the first three hidden neurons.)
depending on whether the nonlinear term is included (A) or not (D) in SAL
when SAL was applied to an RNN together with BPTT.
Each of them is an example in the case of Fig.~\ref{fig:ComparisonRNN} (A) or (D),
and the initial weights are the same as in Fig.~\ref{fig:Learning_RNN}.
}
\label{fig:Seq_comp_linR}
\end{figure*}
The result for the combination of `linear SAL' and `nonlinear criterion' is shown
in Fig.~\ref{fig:ComparisonRNN}(D), but learning succeeded only in one simulation run.
To investigate the failures, a sample learning process for the 50 epochs from the beginning
is shown in Fig.~\ref{fig:Seq_comp_linR}.
The initial weights were the same as the case of Fig.~\ref{fig:Learning_RNN}.
In the `original' case (A), as shown in Fig.~\ref{fig:Seq_comp_linR}(A)-(a) and (c),
once the sensitivity was greater than 1.0,
the weight vector did not increase anymore.
The mean absolute value of the output was almost less than 0.5
as in Fig.~\ref{fig:Seq_comp_linR}{(A)-(b)}.
On the other hand, in the case of linear SAL (D),
as in Fig.~\ref{fig:Seq_comp_linR}(D)-(a),
the sensitivities dropped from around 1.0 in the 2nd epoch in two of the three hidden neurons
(the first (blue) and third (green) lines) even though their weight size continued to increase
as in Fig.~\ref{fig:Seq_comp_linR}(D)-(c).
At that time, the mean absolute outputs of the corresponding neurons in (D)-(b)
also increased even with a large fluctuation due to the presented pattern.
That means that the decrease in $f'(U)$ due to the output increase towards the saturation region
caused the sensitivities to decrease.
Therefore, it is suggested that the nonlinear term in SAL worked to 
keep the sensitivities high by avoiding the outputs from entering the saturation area 
and made the difference in performance between the cases (A) and (D).

In the case (E) when SAL and criterion were both linear, 
learning was successful in half of the runs as shown in Fig.~\ref{fig:ComparisonRNN}(E).
Since the linear criterion does not include $f'(\cdot)$,
the linear SAL stopped before the weights became very large.
The success ratio is only slightly greater than the best ratio
when the scale of the weight matrix is manually explored in Fig.~\ref{fig:ComparisonRNN}(G).
That suggests that the nonlinear property is the origin of the positive influence of SAL
on BPTT learning.

\subsubsection{Case of Deep Feedforward Neural Network (DFNN)}
Finally, sensitivity adjustment learning (SAL) is applied to a DFNN.
To see the effect of the number of layers clearly, an 8-bit parity problem
with noise addition is employed as a simple nonlinear problem.
There are eight inputs, each of them takes a value of $-1$ or $1$.
There is only one output, and if the number of 1 values in the inputs is odd, the ideal value is $0.8$,
and $-0.8$ otherwise.
The number of patterns is $2^8=256$.
Number of layers is varied, and each layer has 20 neurons.

Table~\ref{table:parameters_FNN} presents the used parameters.
Except for the bottom hidden layer that receives inputs, the learning rate $\eta_{BP}$ for SGD
is the same for all the layers but different depending on the number of layers
as shown in Table~\ref{table:learning_rate_FNN}.
Initial connection weights are decided randomly in the range of $[-init\_W, init\_W]$.
Here, $init\_W$ is called `initial weight scale' that is also the same for all the hidden layers.
The initial values and learning rate for the biases were decided in the same way
as the weights in the same layer.
\begin{table}[htb]
  \begin{center}
    \caption{Parameters for supervised learning of 8-bit parity problem with noise addition
    using SAL and BP in a DFNN.}
    \vspace{3mm}
      \begin{tabular}{c|c||c} \hline
        \multicolumn{2}{c||}{Number of neurons (input,  \ldots , output)}&(8, 20, \ldots, 20, 1)\\ \hline 
	\renewcommand{\arraystretch}{0.9}
        \multirow{4}{*}{\begin{tabular}{c}Initial \\ connection weights\\ \small{(uniformly random)} \end{tabular}} 
	\renewcommand{\arraystretch}{1.0}
       & input $\rightarrow$ hidden & \multirow{2}{*}{$[-0.1, 0.1]$} \\
        & hidden $\rightarrow$ output & \\ \cline{2-3}
        & \multirow{2}{*}{hidden $\rightarrow$ hidden} 
	& \multirow{2}{*}{\begin{tabular}{c}varied or \\ $[-0.1, 0.1]$ \end{tabular}} \\& & \\ \hline
        \multirow{2}{*}{\begin{tabular}{c}Learning rate $\eta_{BP}$ \\ in Eq.~(\ref{Eq:Dw_BP}) \end{tabular}}
        & \multirow{2}{*}{input $\rightarrow$ hidden} & \multirow{2}{*}{$0.02$} \\ & & \\ \hline
        \multicolumn{2}{c||}{Learning rate $\eta_{SAL}$ in Eq.~(\ref{Eq:Delta_w})}&0.001\\ \hline
        \multicolumn{2}{c||}{Decay rate $\beta$ in Eq.~(\ref{Eq:decay})}&0.99\\ \hline
        \multicolumn{2}{c||}{Noise added to each input (uniformly random)}&[-0.2, 0.2]\\ \hline
        \multicolumn{2}{c||}{Learning epochs}&5000\\ \hline
       \end{tabular}
    \label{table:parameters_FNN}
  \vspace{6mm}
  \caption{Learning rate $\eta_{BP}$ in Eq.~(\ref{Eq:Dw_BP}) for the weights other than
  those from the inputs to the hidden neurons.
  It depends on the number of layers of the used DFNN.}
  \vspace{3mm}
  \begin{tabular}{c||c|c|c|c|c|c|c|c}\hline
    Number of layers & 3 & 5 & 10 & 30 & 100 & 200 & 300 & 1000\\ \hline
    Learning rate & 0.01 & 0.003 & 0.001 & 0.0007 & 0.0005 & 0.0004 & 0.0003 & 0.0001\\ \hline
  \end{tabular}
    \label{table:learning_rate_FNN}
  \end{center}
\end{table}

\begin{figure}[pthb]
\centerline{\includegraphics[scale=0.65]{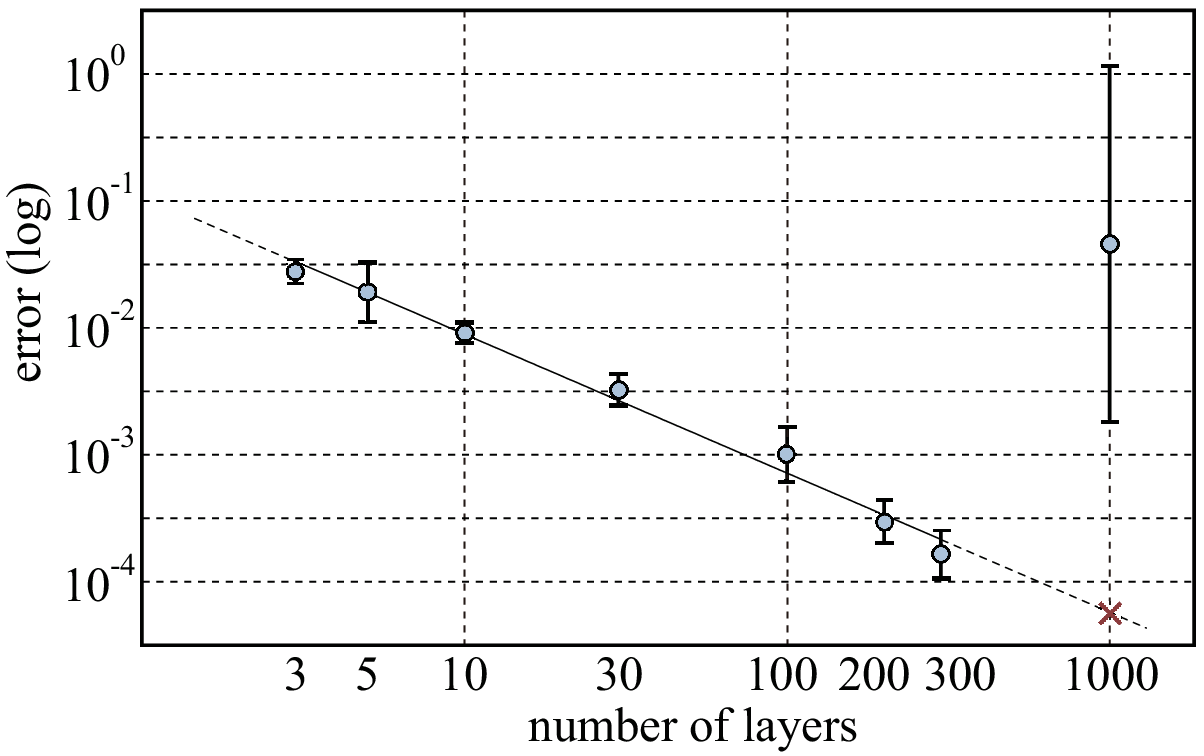}} 
\caption{The relation between the number of layers and RMS error over 256 patterns in log-scale
after learning when SAL was applied with BP in a DFNN with an initial weight scale of 0.1.
The average and standard deviation over 20 simulation runs are shown for each plot.
The '${\textbf{$\times$}}$' mark at the 1000 layers shows the error for only one successful run.
The line indicates the linear approximation for the range from 3 layers to 300 layers.}
\label{fig:Layer_Error}
\vspace{6mm}
\begin{center}
\includegraphics[scale=0.67]{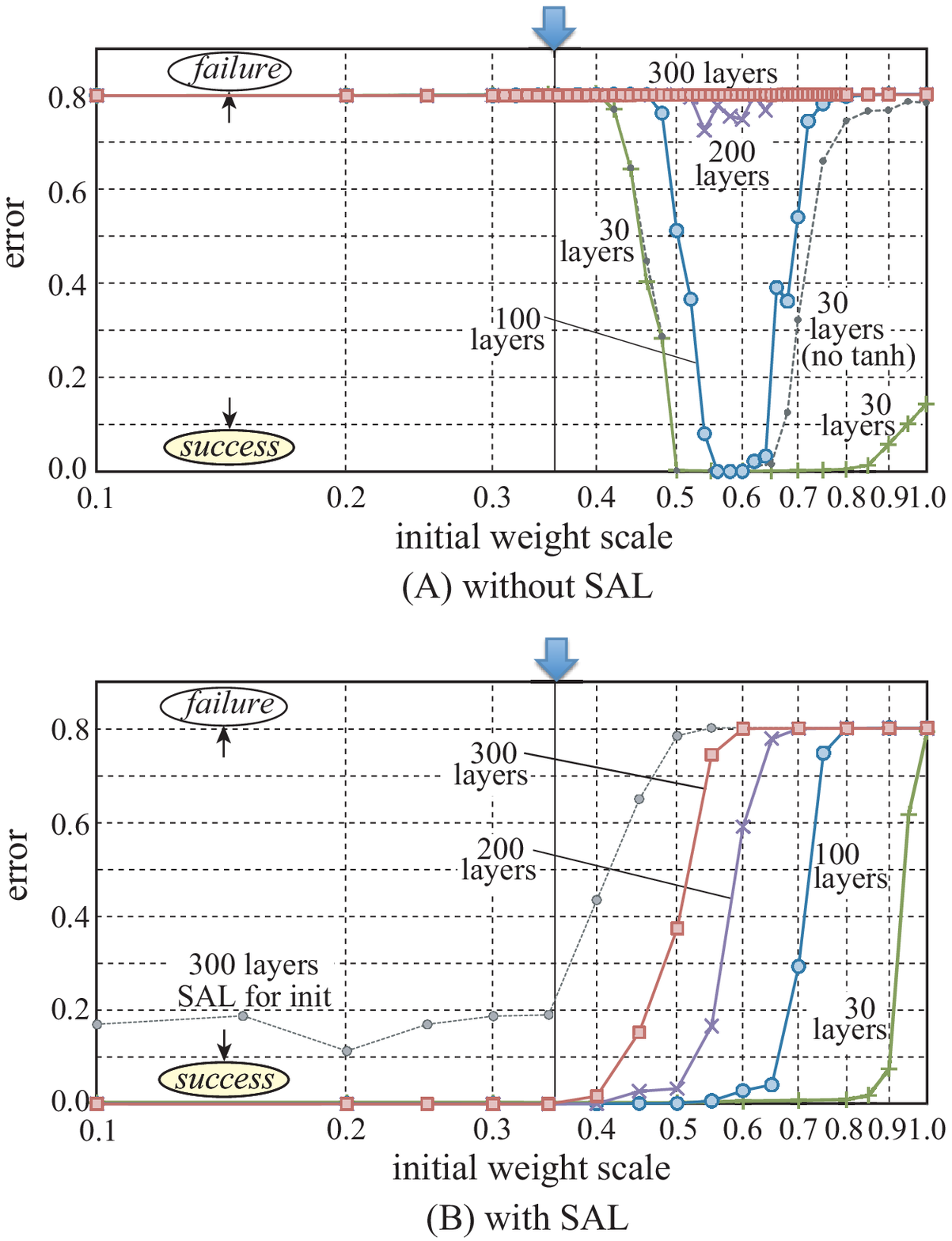}
\end{center}
\caption{Comparison of the error after supervised learning for various initial weight scales
in a DFNN between the two cases of (A) only BP (`without SAL') and (B) SAL+BP (`with SAL').
The number of layers was varied in 30, 100, 200, 300.
Each plot shows the RMS error for 256 patterns averaged over 20 simulation runs.
In (A), the case without $\tanh$ in BP is also shown only for the case of 30 layers.
In (B), the case of applying SAL only until the sensitivity reaches 1.0 for the first time
is also shown for the case of 300 layers.
The vertical line and the arrow around 0.36 on the horizontal axis indicate
that the expected spectral radius of the weight matrix is 1.0.}
\label{fig:init_w-error}
\end{figure}

At first, learning performance is observed by changing the number of layers from 3 to 1000.
Fig.~\ref{fig:Layer_Error} shows the mean and standard deviation
of log-scaled RMS error over 256 patterns after learning in 20 simulation runs
with a different random sequence for initial weights and noises.
The NN can learn the problem even with 300 hidden layers without employing special architectures.
It can be seen that the performance is improved as the number of layers increases until 300 layers,
and the errors lie almost on the linear approximation in the log-scale.
In the case of 1000 layers, learning succeeded only once.
However, the error was the smallest of all the $20 \times 8$ runs and lay
on the linear approximation for the range from 3 to 300 layers.

Next, learning performances with various initial weight scales are observed.
Fig.~\ref{fig:init_w-error}(A) shows the results when SAL was not applied,
and Fig.~\ref{fig:init_w-error}(B) shows the results when SAL was applied.
We call them `without SAL' and `with SAL', respectively.
The errors are plotted for four cases varying the number of layers as 30, 100, 200, 300,
and each plot is the average error over 20 simulation runs with different random sequences.

In Fig.~\ref{fig:init_w-error}(A) for `without SAL',
when the initial weight scale was less than 0.4,
learning failed even in the case of 30 layers.
The learning performance was the best around the initial weight scale from 0.55 to 0.6,
where the expected spectral radius (maximum absolute eigenvalue)
of the weight matrix is around 1.6.
The error around there is lower as the number of layers is smaller.
When the weight scale becomes greater than the optimal one, the error increases again.
Those can be due to the influence of the vanishing/exploding gradient.
However, when the number of layers is 300, even though the step size of varying
the initial weight scale is as small as 0.01, there is no scale with which the error
becomes less than around 0.8.

Fig.~\ref{fig:init_w-error}(B) shows the result when SAL is applied with BP.
When the initial weight scale was large, learning failed.
However, it can be seen that when the initial weight scale was set to a small value like 0.1,
learning succeeded in all the 20 simulation runs regardless of the number of layers.
The boundary scale between learning success and failure is smaller as the number of layers is greater.
From Fig.~\ref{fig:Layer_Error} where the vertical axis is log-scaled,
the error is smaller as the number of layers is larger when the weight scale is 0.1.
That is the opposite trend of the case of `without SAL'.

\begin{figure}[htb]
\centerline{\includegraphics[scale=0.67]{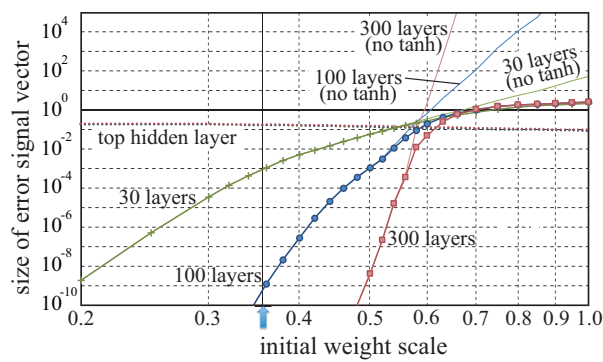}}
\caption{RMS of the propagated error signal $\delta$ at the bottom hidden layer before learning
for the cases of 30, 100, 300 layers.
For comparison, the error signal at the top hidden layer is also shown by three dotted lines,
but because of the overlap of them, it is difficult to find two hidden lines.
}
\label{fig:init_w-delta}
\end{figure}

Fig.~\ref{fig:init_w-delta} shows the size of the propagated error signal $\delta$
at the bottom hidden layer before learning varying the initial weight scale
for the cases of 30, 100, 300 layers.
That at the top hidden layer is also plotted for comparison
and is around $10^{-1}$ not depending so much on the number of layers or the initial weight scale.
At the arrow around the weight scale of 0.36, the spectral radius of each weight matrix
between hidden layers is expected to be 1.0.
However, when the scale is around 0.58, the error signal vector has almost the same size
between the top and bottom hidden layers.
The weight scale is almost the same as when the performance is the best in Fig.~\ref{fig:init_w-error}(A).
That suggests the vanishing/exploding gradient made the learning difficult
when SAL was not applied.

On the other hand, when the number of layers was 300, the error never went down for any weight scale
as in Fig.~\ref{fig:init_w-error}(A) even though the error signal reached the bottom layer
for the weight scale around 0.6.
In the case of 200 layers, the error becomes slightly small around initial weight scale 0.58,
but the error was considerably smaller when SAL was applied with a small initial weight scale
as in Fig.~\ref{fig:init_w-error}(B).
That implies that simply adjusting the initial weight scale or spectral radius is insufficient to learn appropriately.

We consider two reasons for the excellent performance when SAL was applied.
The first one is the effect of continuous application of SAL.
The error when applying SAL only at the beginning of learning, is additionally plotted
in Fig.~\ref{fig:init_w-error}(B) for the case of 300 layers.
The learning sometimes failed as well as the case of RNN in Fig.~\ref{fig:ComparisonRNN} (B).
That means that the sensitivity needs to be around its moderate level
not only at the beginning of learning but also during learning.
However, in this case, although it occasionally failed to learn, in many other  runs, 
the error was equivalent to the case when SAL is always applied.
The second reason could be the limitation of adjusting the spectral radius of the weight matrix.
Even though the spectral radius is appropriate, the sensitivities may not be appropriate
for all the neurons, especially when the number of inputs is small.
In contrast, SAL makes them moderate in each neuron through learning.
Actually, it was confirmed that when the Euclidean norm of the weight vector $|{\mathbf w}|$
was adjusted around 1.2 individually in each neuron before learning,
the average RMS error went down around 0.75
even without applying SAL in a 300 layer DFNN.

\begin{figure}[htb]
\centerline{\includegraphics[scale=0.4]{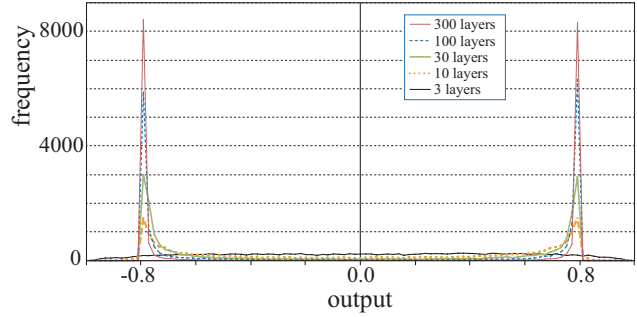}} 
\caption{Histogram of the network output when 1000 continuous random input vectors
whose element was decided randomly from $-1.0$ to $1.0$ was given to 20 networks after learning
of 8~bit parity problem with noise addition.
The number of layers is varied from 3 to 300.}
\label{fig:Histogram}
\end{figure}

Finally, processing through layers is investigated
by observing the output when continuous random inputs were given
to the 20 NNs trained with different initial weights.
1000 sets of 8 continuous uniform random numbers ranging from $-1.0$ to $1.0$ were used as input,
and the output was observed for the total of $1000\times20=20000$ cases.
Fig.~\ref{fig:Histogram} shows the histogram of the output when the number of layers
was varied in 3, 10, 30, 100, 300.
As the number of layers increases, the frequency becomes larger for the output being
around $-0.8$ or $0.8$.
Attractor-like processing through layers can be seen when the number of layers is large.
We think that brought out the high learning performance for the noisy inputs
as presented in Fig.~\ref{fig:Layer_Error}.

\begin{figure}[htb]
\vspace{7mm}
\centerline{\includegraphics[scale=0.58]{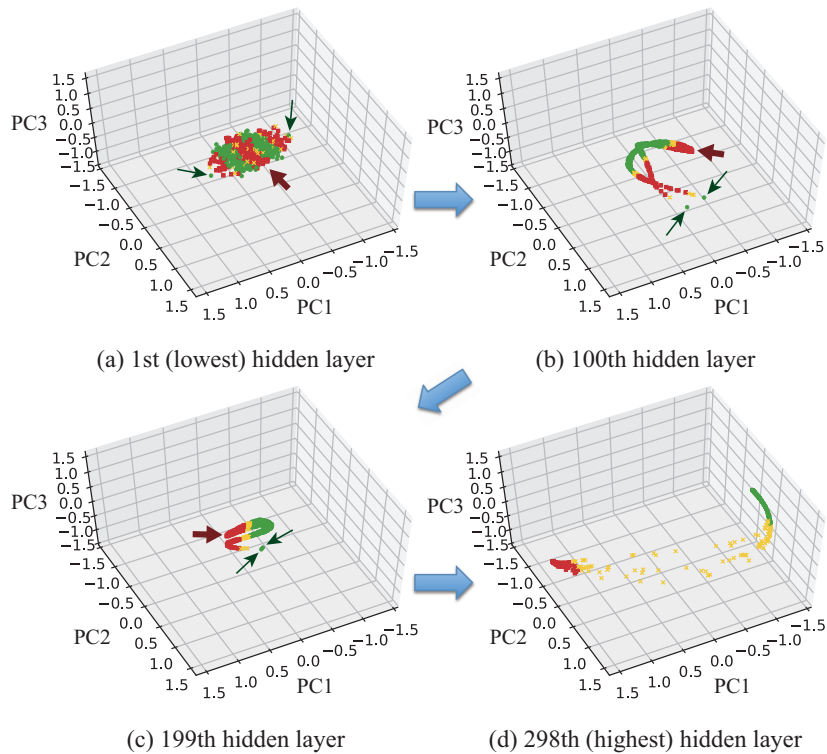}} 
\caption{Change of internal representation after PCA through layers
for 1000 continuous random input patterns after learning.
(One sample of the 20 networks in Fig.~\ref{fig:Histogram})
The plot color and shape indicate the final network output as
'{\color[rgb]{1.0, 0.0, 0.0}{\scriptsize{$\blacksquare$}}}'(red): $0.75<o<0.85$, '{\color[rgb]{0.13,0.55,0.13}{\large{$\bullet$}}}'(green): $-0.85<o<-0.75$, '{\color[rgb]{0.8, 0.64, 0.0}{\textbf{$\times$}}}'(gold): others.
Each of the three arrows shows how a specific point or group of points moves through layers.}
\label{fig:DNN_PCA}
\end{figure}

\vspace{3mm}
Then the transition of the hidden representation is observed when the number of layers is 300.
Fig.~\ref{fig:DNN_PCA} shows the hidden outputs for the random 1000 sets of continuous inputs
after PCA (principal component analysis)
in each of the 1st (bottom), 100th, 199th, 298th (top) hidden layers.
Among 20 simulation runs, one sample that makes it easy to see the transition in representation is picked up.
It can be seen that the representation is extended and folded gradually like the baker's transformation,
and the internal states are divided into two lumps finally.
$\Box \circ$

\section{Discussion \& Conclusion}
We believe that in the future, with the increasing demand for higher functions
such as conversation and thinking,
the control of dynamics in recurrent neural networks (RNNs) will be more and more critical.
Towards such a future, we have shown the possibility that only by controlling the sensitivity in each neuron,
the global dynamics of RNNs can be controlled.
When applying the sensitivity adjustment learning (SAL) to an RNN, the log sensitivity
(for layered RNNs, sum of each layer's log sensitivity)
is almost identical to the maximum Lyapunov exponent until the network dynamics get into chaos
not depending on the number of neurons, connection ratio or number of layers.
In particular, when the sensitivity of each neuron is adjusted to 1.0,
the maximum Lyapunov exponent of the entire network becomes 0.0.
That means the dynamical state of `edge of chaos', which is very important from
an information-processing perspective, is achieved autonomously in an RNN through SAL.

In gradient-based learning, maintaining a small deviation through layers in the forward computation
and maintaining the gradient or error signals in the backward computation
are both represented by using the Jacobian of the layer's computation,
and are equivalent to each other.
Therefore, SAL with 1.0 target sensitivity
leads to avoiding `vanishing gradient' in gradient-based learning in RNNs,
which is also valid in deep feedforward neural networks (DFNNs).
The combination of SAL and BP or BPTT enables to solve the problems
without introducing special architecture or computation
except for the general and local learning from small initial weights.
The performance is better than when the initial weight scale is finely tuned without applying SAL.
When the values of the weights are small, SAL simply increases the magnitude of each weight vector
without changing the vector direction, and so it seems to be equivalent
to just increasing the weights gradually at a glance.
However, the advantage of locality, continuity, and nonlinearity in SAL
can be seen from the above learning results as follows.
\begin{itemize}
\item
The computation is entirely local, and also the decision to stop the SAL can be made
autonomously by each neuron itself.
However, each neuron cannot compute the spectral radius or eigenvalues of the weight matrix
between layers locally.
\vspace{-0.3mm}
\item
Sensitivity is not adjusted for each layer but more finely adjusted for each neuron.
\vspace{-0.3mm}
\item
The sensitivity considers not only linear transformation by the weight vector
but also the nonlinear transformation by the activation function.
Therefore, different from the adjustment of the spectral radius of the weight matrix,
SAL can control the actual dynamics directly considering non-linear processing.
By setting the target value to be 1.0, `edge of chaos' can be realized easily.
\vspace{-0.3mm}
\item
Nonlinear term in SAL prevents the output from going into the saturation area of the activation function.
That enables the network to maintain  good information transmission
in both forward (output) and backward (error) computation.
\vspace{-0.3mm}
\item
Applying SAL with another learning together prevents the loss of sensitivity
caused by the other learning.
\end{itemize}
They give us the hope that SAL will make us free from the fine-tuning of the initial weight
and bring us better learning performance than the manual tuning.
In reservoirs, which have yielded outstanding results in time-series data processing and
temporal pattern generation, an appropriate scale of random weight values is essential.
The proposed method can also be an answer to the question of how each neuron
gets such moderate weights autonomously.

The following are the immediate issues we are currently facing.\\
\vspace{-1.0 \baselineskip}
\setlength{\leftmargini}{18pt}
\begin{description}
\setlength{\itemsep}{-10pt}
\leftskip = 12pt
\item{(1) Application to more complicated problems and comparison with other methods}\\
This paper has focused on the analysis of SAL behavior,
and we employed simple learning problems when examining the combination of SAL and BP or BPTT.
Therefore, it is strongly expected to apply SAL to more practical and large-scale
complicated problems and compare the performance with other methods.\\
\item{(2) SAL for different kinds of input}\\
Another essential unsolved issue is to verify whether SAL works more generally,
especially, when a neuron receives both external signals and those from other neurons.
The case also should be examined in which the RNN's architecture is not simple,
but more complicated with multiple loops.
In reinforcement learning problems, the feedback loop through the outer world
from actions to perceptions for an agent also influences the network dynamics.
In this paper, SAL could control the global dynamics of a two-layer network
even though the neurons in the other layer were considered as an external world during learning. \\
\item{(3) Setting initial connection weights}\\
SAL autonomously adjusts connection weights, but small and random initial weights are still assumed.
For being more plausible, a way to determine the initial weights without using random numbers
should be investigated, for example,
by considering the physical distance to downstream neurons.
The influence of non-random initial weights should be also investigated.\\
\item{(4) Adjustment of learning rate}\\
While we hope SAL reduces the load of tuning the initial weights,
naive learning rate adjustment is still necessary.\\
\item{(5) Conflict to other learning}\\
SAL sometimes increases the error for the other learning like the case in Fig.~\ref{fig:Learning_RNN}(b).
The influence should be investigated in more detail.\\
\item{(6) SAL on dynamical neurons}\\
It also should be examined whether SAL works in dynamical neurons (continuous-time model),
enabling various time constant or chaotic neurons with refractoriness.
\end{description}

SAL is also significant in terms of autonomous, distributed, and asynchronous processing
that enables to control the global dynamics of an RNN by local learning in each neuron.
In the current neural network computation, we often utilize the parallel processing in GPUs,
but they are still under a centralized system.
In the future, if the network becomes larger and more flexible like our brain and
is trained and utilized in real-time, parallelization at the level of each neuron
and autonomous decentralized processing will be essential~[\cite{Moderatism}].
The proposed method in this paper lies in this direction,
and we also expect to accelerate researches in this direction.

We believe that SAL shows its potential by using it together with another learning for its original purposes,
such as reinforcement learning and supervised learning.
Here, we try to generalize the idea about this framework.
In an RNN, the other learning forms attractors to increase reproducibility for better states or actions
and turns the state transitions from irregular to rational.
However, the formation of attractors causes a decrease in sensitivity (chaoticity).
As a result, it is possible that the network is stuck in an attractor and remains inactive for a long time.
That would mean``death'' for the learning agent and should never happen.

If the time average of the sensitivity in each neuron is 1.0, the network dynamics is
Edge of Chaos that lies on the boundary between chaos and non-chaos.
However, this does not mean that the network always keeps the size of a tiny variation of a signal
constant without diverging or converging.
When an attractor is formed by the other learning under the constraint of average sensitivity to be 1.0,
the attractor is not complete but becomes a pseudo-attractor, and both convergence
and divergence appear alternately while being balanced in the course of time.
That enables the agent can balance exploration and learning and maintain autonomous state transitions, 
ensuring the agent does not reach the``death''.
In the high-dimensional space formed by a large number of neurons, pseudo-attractors
would created through learning and chaotic itinerancy among them emerges,
which leads to realizing ``thinking'' including ``inspiration'' and ``discovery'', we expect.

We are living not only in space but also in time.
However, most of the existing learning methods have focused on the state or output at a timing.
We can express it as a point in space, and learning has moved it to a better place.
Even though we use an RNN and its state or output changes over time,
learning has not taken into account its flow or lines in space.
The sensitivity is an index for the flow in the processing of one neuron.
We hope the sensitivity is the key to developing learning methods to control the flow or dynamics.
What we want to do most is to bring this idea of ``learning of dynamics'' into reinforcement learning
and establish a new paradigm.
Learning does not aim to improve the output or state at a specific time
but improve the dynamics from the viewpoint of ``exploration'' or ``reproducibility''
in high-dimensional systems.
We believe that is a fundamental idea towards the ultimate goal: ``emergence of thinking''.